\documentclass[sn-basic]{sn-jnl}

\usepackage{graphicx}%
\usepackage{multirow}%
\usepackage{amsmath,amssymb,amsfonts}%
\usepackage{amsthm}%
\usepackage{pifont}
\usepackage{mathrsfs}%
\usepackage[title]{appendix}%
\usepackage[table,xcdraw]{xcolor}
\usepackage{textcomp}%
\usepackage{hyperref}
\usepackage{bookmark}
\usepackage{manyfoot}%
\usepackage{booktabs}%
\usepackage{algorithm}%
\usepackage{algorithmicx}%
\usepackage{booktabs}\let\cline\cmidrule
\usepackage{algpseudocode}%
\usepackage{multirow}
\usepackage{listings}%
\usepackage{lineno}
\usepackage{here}
\usepackage{stfloats}
\usepackage{anyfontsize}
\usepackage{lmodern}
\usepackage[figuresright]{rotating}
\graphicspath{{figures/}}
\usepackage[normalem]{ulem}
\useunder{\uline}{\ul}{}

\newcommand{\setParDef}{\setlength {\parskip} {0pt} }

\begin{document}

\title[OCCO: LVM-guided Infrared and Visible Image Fusion Framework based on Object-aware and Contextual COntrastive Learning]{OCCO: LVM-guided Infrared and Visible Image Fusion Framework based on Object-aware and Contextual COntrastive Learning}


\author*[1]{\fnm{Hui} \sur{Li}}\email{lihui.cv@jiangnan.edu.cn}

\author[1]{\fnm{Congcong} \sur{Bian}}\email{bociic\_jnu\_cv@163.com}

\author[1]{\fnm{Zeyang} \sur{Zhang}}\email{zzy\_jnu\_cv@163.com}

\author[1]{\fnm{Xiaoning} \sur{Song}}\email{x.song@jiangnan.edu.cn}

\author[2]{\fnm{Xi} \sur{Li}}\email{xilizju@zju.edu.cn}

\author[1]{\fnm{Xiao-Jun} \sur{Wu}}\email{wu\_xiaojun@jiangnan.edu.cn}

\affil[1]{\orgdiv{International Joint Laboratory on Artificial Intelligence of Jiangsu Province}, \orgname{School of Artificial Intelligence and Computer Science, Jiangnan University}, \orgaddress{\street{LihuRoad}, \city{Wuxi}, \postcode{100190}, \state{Jiangsu}, \country{China}}}

\affil[2]{\orgname{College of Computer Science and Technology, Zhejiang University}, \orgaddress{\city{ Hangzhou}, \postcode{310007}, \state{Zhejiang}, \country{China}}}



\abstract{Image fusion is a crucial technique in the field of computer vision, 
and its goal is to generate high-quality fused images and improve the performance of downstream tasks. 
However, existing fusion methods struggle to balance these two factors. Achieving high quality in fused images may result in lower performance in downstream visual tasks, and vice versa.
To address this drawback, a novel LVM (large vision model)-guided fusion framework with \textit{O}bject-aware and \textit{C}ontextual \textit{CO}ntrastive learning is proposed, termed as \textit{OCCO}. 
The pre-trained LVM is utilized to provide semantic guidance, allowing the network to focus solely on fusion tasks while emphasizing learning salient semantic features in form of contrastive learning. Additionally, a novel feature interaction fusion network is also designed to resolve information conflicts in fusion images caused by modality differences. By learning the distinction between positive samples and negative samples in the latent feature space (contextual space), the integrity of target information in fused image is improved, thereby benefiting downstream performance. 
Finally, compared with eight state-of-the-art methods on four datasets, the effectiveness of the proposed method is validated, and exceptional performance is also demonstrated on downstream visual task.}




\maketitle

\section{Introduction}\label{sec1}

The multi-modal image fusion task aims to merge different modal images captured by various sensors in the same scene \citep{LIUsurvy}, and fully leveraging complementary features to generate a fused image containing effective information. 

In infrared and visible image fusion, visible images offer abundant texture details and rich contextual scene information.  However, under adverse conditions, such as darkness or object occlusion, the utility of visible information may be substantially reduced \citep{ivif-survy}.  Conversely, infrared imaging, which senses thermal radiation, possesses the intrinsic advantage of being considerably less susceptible to environmental factors. By leveraging the complementary characteristics with a well-designed fusion strategy, the fused image will contain the rich scene information and highlight salient objects. Theoretically, this will lead to better performance in advanced visual tasks such as tracking \citep{track-fusion}, object detection \citep{detection23,ijcvdetection}, semantic segmentation \citep{cmx}, etc. 

Infrared and visible image fusion can be categorized into three types based on fusion orientation: (1) visual-driven methods \citep{MS-1,li2018densefuse,ma2019fusiongan}, (2) utilizing multi-modal complementarity to directly impact downstream tasks \citep{cmx,rgbtdetection}, and (3) task-driven approaches \citep{seafusion,segmif}. Over time, a multitude of non-deep learning \citep{MS-2,hybird} and deep learning \citep{li2018densefuse,ma2019fusiongan} methods have been created and applied in image fusion, leading to improved image quality. However, infrared and visible image fusion aims to preserve source information as much as possible, while downstream tasks require clear boundaries. This disparity restricts the further advancement of image fusion. 

\begin{figure}[!tbp]
\centering
\setlength{\abovecaptionskip}{0.cm}
\includegraphics[width=1\linewidth]{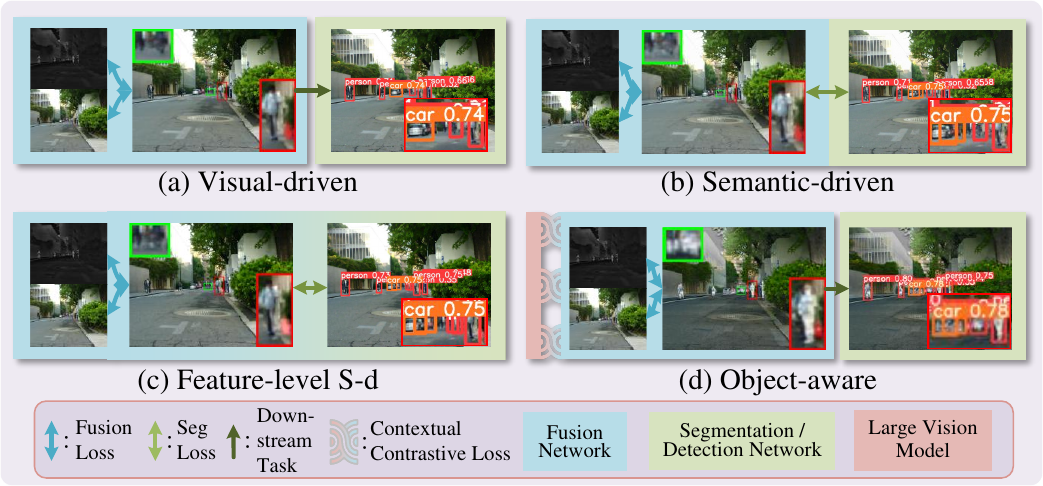}
\caption{From (a) to (d) are visual-driven method, semantic-driven method, feature-level semantic-driven method and object-aware method, respectively. The visual quality in (a) is outstanding with preserved information integrity, albeit showing slightly weaker performance in downstream tasks. Combining segmentation networks in both (b) and (c) enhances downstream performance but has an impact on the quality of the fused images. Focusing exclusively on the fusion task in (d) improves downstream performance without disrupting the fusion results.
}
\label{fig:1}
\end{figure}

Therefore, task-driven methods are crucial for image fusion. Visual-driven methods focus on visual perception and make decisions on information retention or elimination based on algorithms, potentially reducing the importance of essential objects due to low intensity. In Fig.\ref{fig:1}(a) \citep{cdd}, for optimal visual outcomes, uniform information retention without highlighting key scene elements lead to false negative and low confidence in object detection. To address this drawback, some scholars propose training fusion networks with segmentation networks in a cascaded manner to enhance performance in downstream tasks, as shown in Fig.\ref{fig:1}(b) \citep{seafusion}. Although detection results are outstanding, the substantial differences between the two tasks can disrupt fusion outcomes, resulting in fragmented information retention. Moreover, researchers explores methods to effectively interact the features between two tasks to achieve synergistic development. However, for converging with the segmentation loss, the fusion results in Fig.\ref{fig:1}(c) \citep{segmif} exhibit forced information diffusion to enhance edge textures.

To this end, we choose to only utilize the LVM to provide prior semantic information to guide the training, without introducing the segmentation network with segmentation loss, as depicted in Fig.\ref{fig:1}(d). In this way, it not only avoids the gradient disturbance caused by the display constraint, but also retains the semantic perception ability to the greatest extent. 
Contrastive learning optimizes fusion results based on extracted semantic information and enhances semantic content without interfering with the fusion network. Proposed the LVM-guided \textbf{\textit{O}}bject-aware \textbf{\textit{C}}ontextual \textbf{\textit{CO}}ntrastive Learning for infrared and visible image fusion, termed as OCCO, which tackles incomplete semantic information on salient targets due to modality discrepancies. With advancements in LVM, more effective guiding information can be captured. The Segment Anything Model (SAM) \citep{sam}, as a state-of-the-art zero-shot segment large model, it exhibits robust segmentation and generalization capabilities, outperforming other segmentation models. This allows us to thoroughly analyze which information in each modality is more effective for the fused image. Thus, SAM will guide the overall training process as a high-level visual task adviser.

Contrastive learning, as a powerful training strategy in both supervised and unsupervised learning, is particularly well-suited for multi-modal image fusion tasks that lack ground truth. 
Many methods have successfully introduce contrastive learning into multi-modal image fusion tasks \citep{coconet,holoco}. 
Due to the requirement for manual annotation of salient objects, CoCoNet \citep{coconet} is limited to training on the few sample dataset, such as TNO, which lacks scene diversity and texture information in visible images. Thus, in this paper, we propose a novel SAM-guided mask generator, which utilizes a text-driven object detection model Grounding DINO \citep{grounding} to acquire complete detection boxes and used as box prompts in SAM. This processing facilitates the extraction of numerous precisely annotated samples.

The differences in two modalities are visually reflected in the masks. When visible information is rich and infrared information is missing at the same location, the corresponding part will only be displayed in the visible mask. This forms a natural pair of positive samples (rich semantic information) and negative samples (sparse semantic context). We expect the anchor samples (fused image) to contain as much rich information as possible, while staying away from modalities with sparse information. 

In addition, for contrastive learning in multi-modal task with content information, we introduce a new latent feature space, contextual space \citep{cbl}, to calculate the distance between samples. Contextual space is a feature space designed for non-aligned data, comparing the similarity between feature regions based on semantics while ignoring their spatial positions. To deliver more cohesive samples to the contextual space, we designed a feature interaction fusion network. This network strategically prioritizes key features while capturing intricate inter-modality relationships to highlight context-relevant information through the feature interaction fusion block.
In this way, the proposed OCCO solidifies the integrity of salient objects in the fused image, while also preserving the scene information. 

The main contributions of this paper are summarized as follows:
\begin{itemize}

\item A novel LVM-guided contrastive learning training strategy is proposed. It bridges the gap between the fused image quality and the downstream visual task performance, which improves downstream visual task performance without interfering with the fusion results. 

\item The feature interaction fusion network based on the contextual space is designed to enhance the quality of feature representation and improving the sensitivity to spatial context. The key characteristics of contextual is the ability to maintain the appearance of the anchor sample without interference from other factors. 

\item Extensive experiments demonstrate the effectiveness of the proposed fusion network, and it is superior to existing state-of-the-art methods in both fusion performance and downstream performance.
\end{itemize}

The rest of our paper is structured as follows. In Section \ref{sec-relate}, we briefly review the related work on task-driven methods and contrastive learning in deep learning-based image fusion. The proposed SAM-guided object-aware contextual contrastive learning is described in detail in Section \ref{sec-proposed}. The experimental results are presented in Section \ref{sec-experiments}. Finally, we draw the paper to conclusion in Section \ref{sec-con}.

\section{Related Work}\label{sec-relate}

In this section, several typical works related to our method are presented in different subsections. 
Firstly, deep-learning based fusion methods are introduced. Secondly, we will briefly introduce several task-driven infrared and visible image fusion methods. Thirdly, an overview of contrastive learning in multi-modal image fusion is provided.

\subsection{Deep learning-based image fusion methods}
The introduction of deep learning for image fusion is pioneered by Liu et al. \citep{liu2017multi} in multi-focus image fusion. The original authors also develop a CNN-based network to address the challenge of matching weights between infrared and visible images \citep{liu2017infrared}. 

Subsequently, Li et al. propose an auto-encoder based fusion network \citep{li2018densefuse}, integrating DenseNet into the CNN network to tackle the issue of feature loss during extraction. Auto-encoder based network exhibits exceptional feature extraction capabilities and robust feature reconstruction characteristics. However, the fusion strategy in DenseFuse lacks precision. Hence, the original authors suggests a series of two-stage training methods \citep{crossfuse,li2020nestfuse,rfnnest}. Based on auto-encoder network and squeeze-and-decomposition, an efficient real-time image fusion network is also designed by Zhang et al. \citep{Zhang2021SDNetAV}. Moreover, Xu et al. develop a universal fusion network is developed that can merge diverse fusion tasks by employing a continuous training strategy and formulating a parameter retention loss \citep{xu2020u2fusion}. 

Given that Generative Adversarial Networks (GANs) excel in unsupervised tasks, Ma et al. use GANs to perform image fusion \citep{ma2019fusiongan}. They train a generator to generate fused images that the discriminator cannot distinguish as visible image or generated image. Nevertheless, a single discriminator does not effectively ensure that fused images contain information from both modalities. Therefore, a dual-discriminator-based method is proposed to improve fusion quality \citep{ddcgan}. 

Moreover, Transformers also exhibit potent global perception and information enhancement capabilities. Li et al. combine CNN with Transformers \citep{ivfwsr}, enabling CNN to prioritize visible information while Transformers handle infrared information. Ma et al. introduce SwinTransformer into image fusion tasks and proposed cross-modal information interaction \citep{swinfusion}. Li and Wu assess the correlation and uncorrelation information in different modalities \citep{crossfuse}, designing self-attention and cross-attention fusion modules to optimize fusion results effectively. 


\subsection{Task-driven image fusion methods}
Task-driven fusion methods can be broadly classified into two categories: pixel-level constraints and feature-level constraints. Tang et al. are pioneers in integrating high-level vision tasks with the image fusion \citep{seafusion}. By integrating a semantic segmentation network following the fusion network and utilizing a unified loss to govern both networks, the fusion results in segmentation tasks saw enhancement. Similarly, Zhang et al. appended a detection network post the fusion network \citep{detectfusion}, boosting the performance of fused images in detection tasks. In addition, a meta-learning-based training strategy is designed by Zhao er al. to achieve multi-task interleaved training \citep{metafusion} .

Nevertheless, merely concatenating two networks with distinct loss poses challenges for the two tasks to genuinely complement each other. Therefore, Tang et al. suggest engaging with the two tasks at the feature level \citep{PSFusion}. In this method, shallow features handle the fusion task, while deep features manage the segmentation task, effectively leveraging the attributes of diverse tasks. Liu et al. employed cross-attention to allow features from different tasks to interact with each other \citep{segmif}. Wang et al. uses edge preservation technology to generate accurate decision maps and improve the quality of multi-focus image fusion \citep{ynet}. 

Prior methods aim to integrate segmentation into the fusion network and regulate it via loss functions to achieve continuous or parallel training. However, they disregarded the guiding function of high-level visual tasks in semantic-level. This study employs SAM as guidance, concentrating on the precise labels from the existing optimal segmentation network to generate a fusion image highlighting the most crucial information.

\subsection{Contrastive learning in image fusion}
\setParDef
Contrastive learning is well-suited for unsupervised tasks, making it particularly applicable to image fusion. By defining appropriate positive and negative samples, fusion images are prompted to select advantageous feature information reflected in the output. 

Zhu et al. first introduce contrastive learning into image fusion \citep{clfnet}, comparing saliency and texture details between visible and infrared images in a sliding window fashion to blend the fused image with the texture details of the visible image and salient features of the infrared image. CoCoNet \citep{coconet} uses manually annotated salient targets  from infrared images as foreground and non-salient regions from visible images as background to form positive and negative samples. Liu et al. through the color correction module \citep{holoco}, obtain normal brightness information to bring the fused image closer to normal brightness, avoiding overexposure or underexposure.

Inspired by CoCoNet \citep{coconet}, We will further design a method combining with semantic segmentation. CoCoNet solely extracts salient targets from the infrared modality, overlooking valuable information in visible images. Based on SAM, we extract commonly salient targets, such as people and cars on the road, from both modalities to learn their semantic information. These methods all operate within the Euclidean feature space. While Euclidean space effectively captures pixel and gradient information, it does not converge well for feature information. Therefore, we introduce a novel latent feature space to optimize the efficiency of contrastive learning.

\section{Proposed Method}\label{sec-proposed}
\begin{figure*}[t]
	\centering
	\includegraphics[width=1\linewidth]{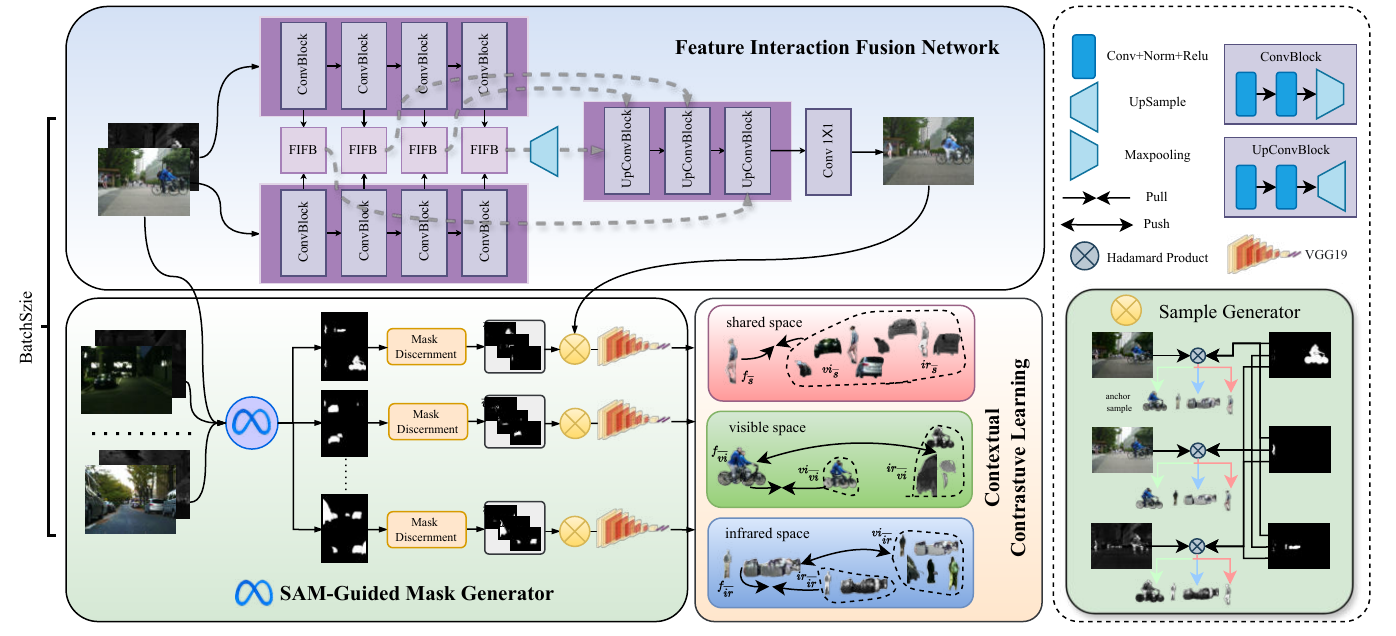}
	\caption{The framework of OCCO and the architecture of Feature Interaction Fusion Network. During the training process, only the first group of fused images within a batch sever as anchor samples, and  first group of source images within a batch treated as positive sample, while all data generate negative samples.}
	\label{fig:pipeline}
\end{figure*}

As shown in Fig.\ref{fig:pipeline}, the framework  of the proposed OCCO is composed of the Feature Interaction Fusion Network (FIFN), the SAM-guided mask generator. Furthermore, a new contextual contrastive learning strategy is also designed to train our framework. This section delves into the network in \ref{network}, mask generator in \ref{Mask}, contextual contrastive learning in \ref{OCCO} and loss function details in \ref{loss}. 

\subsection{Network Architecture }
\label{network}
Inspired by Cross-Modal Feature Rectification Module \citep{cmx}, to gather shallow features and deep semantic features from different modalities, we introduce Feature Interaction Fusion Network (FIFN). 

As illustrated in Fig.\ref{fig:pipeline}, a blue conv unit comprises a $3 \times3$ convolution, a batch normalization, and a ReLU activation function. Furthermore, a conv block includes two conv units and a max-pooling operation, while an upconv block consists of two conv units and an up-sampling operation. After passing the visible image $vi$ and the infrared image $ir$ through four conv blocks with the same structure but different parameters, four single-modal features (${\phi_{vi}}^i\in\mathbb{R}^{B\times H\times W\times C}$, ${\phi_{ir}}^i\in\mathbb{R}^{B\times H\times W\times C}$) can be extracted, where $i \in\{{1,\cdots, 4}\}$ represents the layer number. To ensure that the fused features capture semantic and scene information effectively, Feature Interaction Fusion Block (FIFB) is designed to preserve global spatial information while highlighting modality-specific objects. 

\begin{figure}[t]
	\centering
	\includegraphics[width=1\linewidth]{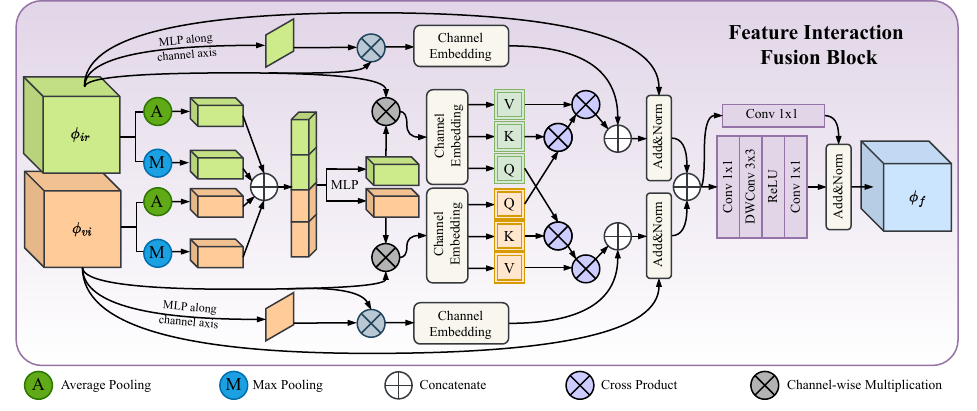}
	\caption{The architecture of FIFB is composed of spatial enhancement, cross channel and cross attention three parts. 
}
	\label{fig:net}
\end{figure}

The structure of FIFB are shown in Fig.\ref{fig:net}. Initially, inter-modal semantic information is enhanced by channel attention. Average pooling preserves the basic information from the global, which belongs to the low-frequency signal. Max pooling, on the other hand, pays more attention to high-frequency details. This combination further ensures the integrity of the information. Concatenate pooling vectors and pass them through an $MLP$ followed by a sigmoid function to obtain channel weights $CW\in\mathbb{R}^{B\times 2\times1 \times C}$. These weights are then split into $CW_{ir}\in\mathbb{R}^{B\times 1\times1 \times C}$ and $CW_{vi}\in\mathbb{R}^{B\times 1\times1 \times C}$. This process can be formalized as follows,
\begin{equation}
    (CW_{vi}^i,CW_{ir}^i)=\mathfrak{S}(MLP(Cat(MA(\phi_{vi}^i),MA(\phi_{ir}^i))),
\end{equation}
where $\mathfrak{S}$ is sigmoid function, $Cat(\cdot,\cdot)$ stands for concatenate operation and $MA$ represent two pooling operations (max pooling and average pooling). $\phi_{vi}^i$ and $\phi_{ir}^i$ are enhanced through the channel attention weights by $(\varrho_{vi}^i,\varrho_{ir}^i)=(CW_{vi}^i\ast\phi_{vi}^i,CW_{ir}^i\ast\phi_{ir}^i)$, where $\ast$ means channel-wise multiplication.
Subsequently, cross-attention mechanism enable the proposed model to learn richer feature representations and enhance the ability to handle complex scenes through the complementary regions of the inter-modality features. $\varrho_{vi}^i$ and $\varrho_{ir}^i$ are flattened into linear sequences inchannel dimension, denoted as $\nu_{vi}^i\in\mathbb{R}^{B\times C\times HW }$ and $\nu_{ir}^i\in\mathbb{R}^{B\times C\times HW}$. Then the sequences are projected into query ($Q_{vi}^i$, $Q_{ir}^i$), key ($K_{vi}^i$, $K_{ir}^i$) and value ($V_{vi}^i$, $V_{ir}^i$). The attention map of $\phi_{vi}^i$ corresponding to $\phi_{ir}^i$ is calculated by crossing $Q_{vi}^i$ and $K_{ir}^i$, the attention map is obtained by $\varpi_{vi\rightarrow ir}^i=Softmax(Q_{vi}^i {K_{ir}^i}^T)$, 
which is used to enhance the mutual part of visible image in the infrared image by  $\varphi_{ir}^i = \varpi_{vi\rightarrow ir}^i V_{ir}^i $. Similarly, $\varphi_{vi}^i$ enhanced by cross attention can be obtained.

Furthermore, the spatial attention is utilized to enhance the salient information of the intra-modality. The spatial attention map is obtained by an MLP along with channel axis. It is followed by spatial-wise multiplication with the input feature. This spatial enhancement process is formulated as
\begin{equation}
    \upsilon_{ir}^i=\phi_{ir}^i \odot Softmax(C_{1\times1}(R(C_{1\times1}(\phi_{ir}^i)))),
\end{equation}
where $\odot$ represents spatial-wise multiplication, $C_{x\times{x}}$ is a convolutional layer with the kernel size $x \times {x}$ and $R(\cdot)$ is relu activation function. $\upsilon_{ir}^i$ and $\upsilon_{vi}^i$ are also flattened into sequences, and concatenate with $\varphi_{ir}^i$, $\varphi_{vi}^i$, respectively. 

Finally, inputs feature $\phi_{ir}^i$ and $\phi_{vi}^i$ are concatenated and embedded into two modality sequences, represented as
\begin{eqnarray}\label{equ:embedded}
\begin{aligned}
    \upsilon_{f}^i =Cat(& \Omega(\phi_{ir}^i+Cat(\varphi_{ir}^i, \upsilon_{ir}^i)), \\
    &\Omega(\phi_{vi}^i+Cat(\varphi_{vi}^i,\upsilon_{vi}^i))),
\end{aligned}
\end{eqnarray}
where $\Omega$ is batch normalization.

For channel restoration and fusion feature refinement, the output feature is calculated by
\begin{equation}
    \phi_f^i = \Omega(C_{1\times1}(R(DC_{3\times3}(C_{1\times1}(\upsilon_{f}^i))))+C_{1\times1}(\upsilon_{f}^i)),
\end{equation}
where $DC_{3\times 3}$ represent depth-wise convolution with $3\times 3$ kernel size. Depth-wise convolution excels at capturing features from each channel by processing them independently, which enables the model to learn more fine-grained features and enhances its overall feature extraction capabilities. 

Additionally, residual connections help stabilize the training phase of proposed model, preventing it from focusing solely on the most prominent information. The four outputs which are extracted by FIFB will be used to reconstruct the fused image.

\subsection{SAM-guided Mask Generator}
\label{Mask}

\begin{figure}[t]
\setlength{\abovecaptionskip}{0.cm}
	\centering
	\includegraphics[width=0.8\linewidth]{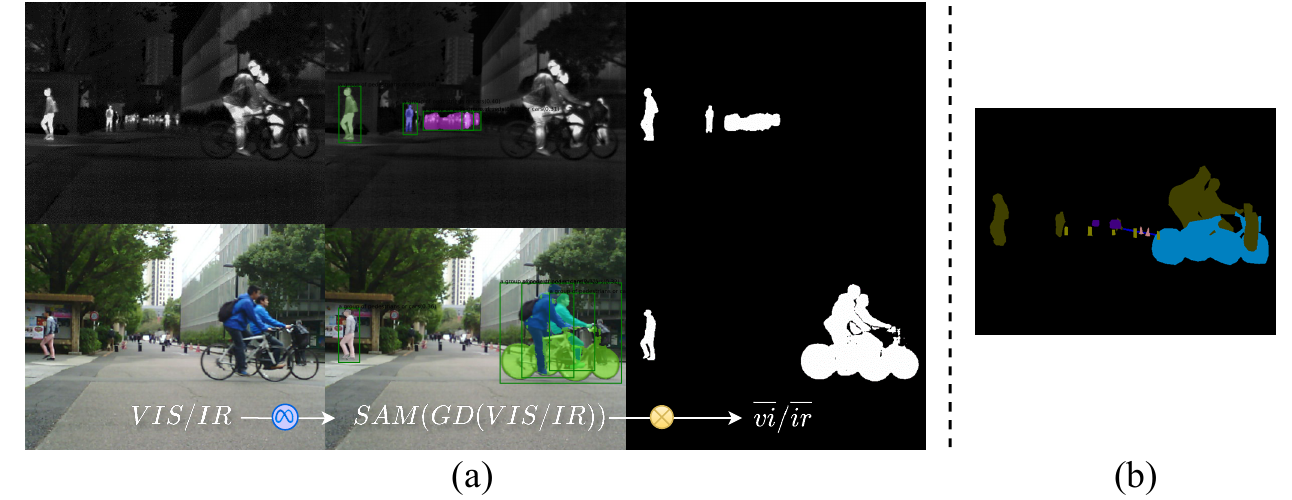}
	\caption{(a) Demonstration of the single-modal mask generation process. (b) Label provided in the MSRS dataset. 
}
	\label{fig:sam}
\end{figure}

SAM can accurately segment objects triggered by various prompts, such as dots or boxes. To enhance accuracy, we employ a text-driven object detection, Grounding DINO \citep{grounding}, to generate detection boxes for SAM. The text prompt ``\textit{The scene captures a group of pedestrians or cars.}'' ensures relevant aspects are captured across both modalities, resulting in two masks:  $\overline{vi} = SAM(GD(vi))$,  $\overline{ir} = SAM(GD(ir))$, where $GD$ indicates the Grounding DINO.

As illustrated in Fig.\ref{fig:sam}, Grounding DINO detect the distant crowd in the infrared image that is not attended to in the label. With the box prompt, SAM accurately segment individuals within the box, offering finer segmentation than the label. These masks convey the utmost semantic data extracted from source images by SAM. 

\begin{figure}[ht]
\setlength{\abovecaptionskip}{0.cm}
	\centering
\includegraphics[width=0.75\linewidth]{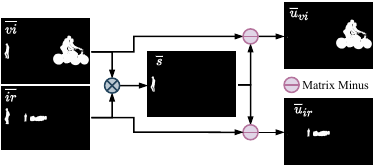}
	\caption{Demonstration of mask discernment. 
}
	\label{fig:mask}
\end{figure}

For the mask information, it is easy to differentiate the two masks into modality-unique information and shared information as shown in Fig.\ref{fig:mask}. The shared information can be derived by intersecting the two masks:  $\overline{ s }=\overline{ vi } * \overline{ ir }$, which $*$ represents the Hadamard product. Modality-unique data is obtained by removing shared information from the original masks: $\overline{ u }_{ vi }=\overline{vi}-\overline{ s}$, $\overline{ u }_{i r}=\overline{ir }-\overline{ s }$. Finally, the background region is obtained by removing these three semantic parts: $\overline{bg} = 1_{H \times W}
- \overline{s}-\overline{ u }_{ vi }-\overline{ u }_{ ir }$, here $1_{H \times W}$ means $H\times W$-shaped all-ones matrix. Thus, we acquire three semantic masks $\overline{s}$, $\overline{u}_{ir}$, $\overline{u}_{vi}$ and a background mask $\overline{bg}$ based on the two source images.

\subsection{Contextual Contrastive Learning}
\label{OCCO}

For contextual contrastive learning, the positive and negative samples are generated by the calculated three semantic masks.

It is apparent that when utilizing $\overline{u }_{ vi }$ as an information filter, the content extracted from the visible image is most beneficial to the segmentation network, while the information from the infrared image is superfluous. Hence, we designate the samples obtained based on $\overline{ u }_{ vi }$ as positive samples from visible images, denoted as $ {vi}_{\overline{vi}  } = { vi } * \overline{u }_{vi }$, and the samples derived from infrared images as negative samples, indicated as $ {ir}_{\overline{vi}  } = { ir } * \overline{u }_{vi }$. The anchor samples at this point are ${f}_{\overline{v i}}={f} *  \overline{u }_{vi }$, where $f$ signifies the fused image obtained through the fusion network. 

Similarly, when generating anchor samples based on the infrared-unique filter from fused images $f_{\overline{ir}}={f} *  \overline{u }_{ir }$, the positive samples and the negative samples can be calculated by $ {ir}_{\overline{ir}  } = { ir } * \overline{u }_{ir }$ and $ {vi}_{\overline{ir}  } = { vi } * \overline{u }_{ir }$, respectively. When shared information as an information filter indicates that both modalities are important, the anchor samples are represented as $f_{\overline{s}}=f * \overline{s}$. Following the previous steps, we perform a contrastive learning process using as ${vi}_{\overline{s}}={vi} * \overline{s}$ positive samples and ${ir}_{\overline{s}}={ir} * \overline{s}$ as negative samples, and then reverse the roles for another round of contrastive learning. 

In summary, contrastive learning will be performed three times based on three semantic information filter. As for the background part, the texture information in the visible image need to be preserved to make the fused image more nature, which means the visible image is positive sample and the infrared image is negative sample.

Having a large number of negative samples from other images benefits anchor sample learning latent features between positive samples and negative samples \citep{crossimagecl}. Hence, we partition the samples equally , resulting in $b$ groups in each batch. Each batch consists of one anchor sample, one positive sample, and $b$ negative samples. This grouping strategy implies that the first group will generate anchor, positive, and negative sample simultaneously, while the remaining  $b-1$ groups only generate negative samples. The contextual contrastive loss is formulated as follows, 

\begin{figure}[t]
	\centering
	\includegraphics[width=0.6\linewidth]{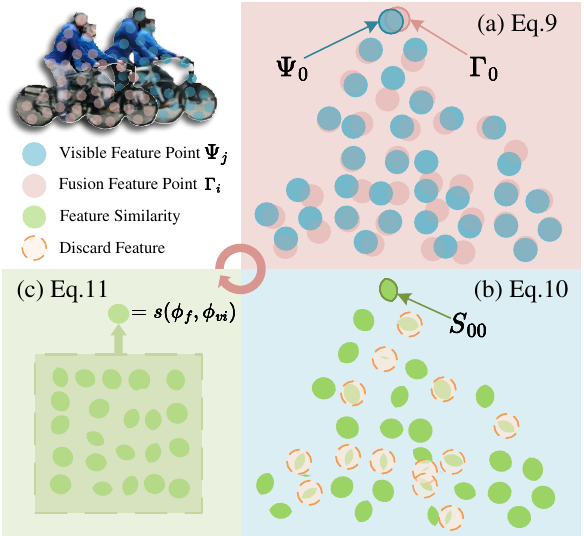}
	\caption{Contextual similarity calculation procedure for fusion and visible features. (a) Equation 9 demonstrates that a larger intersection area indicates higher similarity between two feature points. (b) Equation 10 demonstrates selecting the most similar feature points, which is represented here by discarding dissimilar points. (c) Equation 11 demonstrates using the average value of multiple feature points to represent features' contextual similarity.
}
	\label{fig:cl}
\end{figure}

\begin{equation}
   { L_ {CCL}=L_{uniqe }+L_ {share } +L_ {bg }},
\end{equation}
\begin{equation}
   { L_{ {uniqe }}=\frac{\mathcal{D}\left(f_{\overline{v i}}, v i_{\overline{v i}}\right)}{\sum_{j}^{b} \mathcal{D}\left(f_{\overline{v i}}, i r_{\overline{v i}}{ }^{j}\right)}+\frac{\mathcal{D}\left(f_{\overline{i r}}, i r_{\overline{i r}}\right)}{\sum_{j}^{b} \mathcal{D}\left(f_{\overline{i r}}, v i_{\overline{i r}}{ }^{j}\right)}},
\end{equation}
\begin{equation}
\label{balance}
   { L_{ {share }}=\omega_{1}\frac{\mathcal{D}\left(f_{\overline{s}}, v i_{\overline{s}}\right)}{\sum_{j}^{b} \mathcal{D}\left(f_{\overline{s}}, ir_{\overline{s}}{ }^{j}\right)}+\omega_{2}\frac{\mathcal{D}\left(f_{\overline{s}}, i r_{\overline{s}}\right)}{\sum_{j}^{b} \mathcal{D}\left(f_{\overline{s}}, v i_{\overline{s}}{ }^{j}\right)}  },
\end{equation}
\begin{equation}
     {L_{ {bg }}=\frac{\mathcal{D}_{bg}\left(f_{\overline{bg}}, v i_{\overline{bg}}\right)}{\sum_{j}^{b} \mathcal{D}_{bg}\left(f_{\overline{bg}}, ir_{\overline{bg}}{ }^{j}\right)}},
\end{equation}
where $\mathcal{D}($ · , · $)$ and $\mathcal{D}_{bg}($ · , · $)$ stand for the distance of two samples, which will be introduced in the next paragraph. $\omega_{1}$ and $\omega_{2}$ are fixed weights, which balance the inherent bias in shared information. 

For the latent feature space, CoCoNet selects the commonly used VGG19 model \citep{vgg19} to extract features from five layers and calculate the Euclidean distance between samples. However, samples restricted only with the $l_2$ norm will lack high-frequency information, leading to excessively smooth textures, which is not desirable \citep{cbl}, especially in image fusion task.

In this work, we also leverage the feature extraction capabilities of a pre-trained VGG19 network. However, in contrast to approaches that calculate contextual similarity across all feature layers, we strategically compute it exclusively on deep feature maps, which are known to encapsulate richer semantic information. The contextual loss \citep{mechrez2018contextual}, operates by transforming input feature maps into sets of representative feature points. Specifically, the feature map of the fused image, denoted as $\phi_{f}$, is decomposed into a set of feature points $\Gamma_{i} \in \{\Gamma_{0}, \cdots, \Gamma_{N_f}\}$, as illustrated in Fig.~\ref{fig:cl}. Similarly, the feature map of the visible image, $\phi_{vi}$, is represented by a set of feature points $\Psi_{j} \in \{\Psi_{0}, \cdots, \Psi_{N_{vi}}\}$. Here, $N_{f}$ and $N_{vi}$ represent the number of feature points extracted from $\phi_{f}$ and $\phi_{vi}$, respectively.

Unlike traditional metrics such as Euclidean distance that rely on pixel-wise or feature-wise correspondences, contextual similarity assesses the proximity of semantic relationships between two sets of feature points. The core idea is to measure whether the relative spatial arrangements of feature points in one set are similar to those in another, without requiring a strict one-to-one mapping. To achieve this, we first calculate the contextual similarity $S_{ij}$ between each pair of feature points $\Gamma_{i}$ and $\Psi_{j}$ using cosine similarity. The formulation is given by:
\begin{equation}
\label{eq:s}
S_{ij} = S(\Gamma_{i}, \Psi_{j}) = 1 - \frac{(\Gamma_{i} - \mu_{\Psi}) \cdot (\Psi_{j} - \mu_{\Psi})}{\|\Gamma_{i} - \mu_{\Psi}\|_{2} \|\Psi_{j} - \mu_{\Psi}\|_{2}},
\end{equation}
where $\mu_{\Psi} = \frac{1}{N_{vi}} \sum_{j=1}^{N_{vi}} \Psi_{j}$ represents the mean vector of the feature points $\Psi$, and $\|\cdot\|_2$ denotes the $l_2$ norm. The term $(\Gamma_{i} - \mu_{\Psi})$ effectively centers each feature point $\Gamma_{i}$ with respect to the mean of $\Psi$, enabling the cosine similarity to focus on the relative orientation of feature vectors.

To aggregate the pairwise similarities $S_{ij}$ and obtain a similarity value $s_i$ for each feature point $\Gamma_i$, we employ a normalization process based on the contextual similarities. This process effectively identifies, for each $\Gamma_i$, its most contextually similar counterpart within the set $\Psi$. The similarity value $s_i$ is then computed as:
\begin{equation}
s_i = \max_{k} \left( \frac{\exp(1 - S_{ik})}{\sum_{j=1}^{N_{vi}} \exp(1 - S_{ij})} \right).
\end{equation}
This normalization step highlights the maximum similarity while considering the distribution of similarities across all feature points in $\Psi$.

Finally, to quantify the overall feature-level similarity between $\phi_{f}$ and $\phi_{vi}$, we average the individual feature point similarities $s_i$ across all feature points in $\Gamma$. This yields the feature-level contextual similarity $s(\phi_{f}, \phi_{vi})$:
\begin{equation}
s(\phi_{f}, \phi_{vi}) = \frac{1}{N_{f}} \sum_{i=1}^{N_{f}} s_i.
\end{equation}
This value represents the feature-level constraint imposed by the contextual similarity measure.

While Contextual Loss has been effectively used in style transfer to capture and maintain the relative relationships between features, image fusion tasks necessitate preserving not only relative but also absolute feature characteristics. In image fusion, maintaining the intensity and structural integrity of the source images is crucial. Therefore, we hypothesize that when feature points are contextually similar, their Euclidean distance in feature space should also be minimized to preserve absolute feature properties. To incorporate this, our Contextual Similarity measure is formulated as:
\begin{equation}
{CS}(\phi_{1}, \phi_{2}) = -\log\left(s(\phi_{1}, \phi_{2}) + \omega_{3} e(\phi_{1}, \phi_{2})\right),
\end{equation}
where $e(\phi_{1}, \phi_{2}) = \|\phi_{1} - \phi_{2}\|_{2}$ represents the Euclidean distance between the feature maps $\phi_{1}$ and $\phi_{2}$, and $\omega_{3}$ is a weighting factor that balances the contribution of contextual similarity and Euclidean distance. By incorporating the $l_2$ norm-based Euclidean distance term, we encourage the model to learn not only the semantic arrangement of features but also to preserve essential spatial distribution and texture details, which are vital for effective image fusion.

The spatial gap between different modalities is much bigger on shallow features, forcing them closer on Euclidean distance will lead the anchor samples to learn only high intensity areas and loss the ability to learn the semantics knowledge. Therefore, we focus on deep features in the contextual space. In general, the distance between anchor samples, positive samples and negative samples is designed in this paper is given as follows,
\begin{equation}
{\mathcal{D}\left(f_{\overline{v i}}, v i_{\overline{v i}}\right)}=\sum_{i=4}^{M}CS(v g g_{i}(f_{\overline{v i}}),v g g_{i}(v i_{\overline{v i}})),
\end{equation}
where $v g g_{i}(f_{\overline{v i}})$ means the $i$-th layer features extracted from the VGG19 of the image $f_{\overline{v i}}$ and $M$ represents the maximum number of layers extracted from the vgg19, which is 5 in this paper. We select 4-$th$ layer and 5-$th$ layer features as deep semantic feature. 

Finally, the background information does not contain special semantic information, the European distance is utilized to measure the similarity on shallow feature space ($i=1$),
\begin{equation}
{\mathcal{D}_{bg}\left(f_{\overline{bg}}, v i_{\overline{bg}}\right)}=e(v g g_{1}(f_{\overline{bg}}),v g g_{1}(v i_{\overline{bg}})).
\end{equation}

\subsection{Loss Function}
\label{loss}
Salient objects have been highlighted in the fused image via contextual contrastive loss, while the entire fused image must also adhere to pixel-level loss constraints.
 At first, for structural similarity, we use $SSIM$ metric \citep{ssim} to constrain the brightness, contrast and structural details for both images. It can be represented by the following equation,
\begin{equation}
    L_{ssim}=2-SSIM(f,vi)-SSIM(f,ir).
\end{equation}

For intensity, refer to \citep{sdcfusion},  we aim to integrate high-intensity information in the image. Visible images contain not only high-intensity information but also variations. Thus, $l_2$ norm is used to manage the disparity between the fused image and the visible image. Regarding infrared image, we focus solely on learning the salient regions where infrared brightness surpasses that of the visible image, marked by a simple mask,
\begin{equation}
   { M(i, j)=\left\{\begin{array}{ll}1, & \text { if } I_{i r}(i, j)>I_{v i}(i, j) \\0, & \text { otherwise }\end{array}\right.},
\end{equation}
where $I_{v i}(i, j)$ represents the intensity of the pixel at position $(i,j)$ in the visible image. 

The intensity loss calculation is demonstrated by the equation,
\begin{equation}
    L_{int}=\frac{1}{H W} \sum_{j=1}^{H} \sum_{i=1}^{W}\left(\left\|f-vi\right\|_{2}^{2}+M\left\|f-ir\right\|_{1}\right),
\end{equation}
where $\|\cdot\|_1$ is $l_1$ norm. Due to the modal characteristics, where the visible information is prominent, we not only want it to be enchanced, but also want to retain its continuous gradient information. Thus, $l_2$ norm is utilized to constrain this part. However, for the infrared modality, there is no gradient change inside in most scenario, the $l_1$ norm is used to limit infrared part.

In order to capture rich texture information from input images, the texture loss is indispensable,
\begin{equation}
   {L}_ {texture }=\frac{1}{H W}\left\|\left|\nabla f\right|-max \left(\left|\nabla vi\right|,\left|\nabla ir\right|\right)\right\|_{1},
\end{equation}
where $\nabla$ means refers to the sobel gradient operator, $\mid\cdot\mid$ denotes the absolute operation and $max(\cdot,\cdot)$ stands for the element-wise maximum. 

Thus, the overall pixel loss of this method can be represented as follows,
\begin{equation}
    L_{pixel}= L_{ssim}+\omega_4L_{int}+\omega_5L_{texture},
\end{equation}
where $\omega_4$  and $\omega_5$ two weights are used to ensure that all three of losses are of the same magnitude. 

Combined with the contextual contrastive loss proposed in this paper, the overall loss function is represented as follows,
\begin{equation}
    L_{total}=L_{pixel}+\omega_6L_{con},
\end{equation}
where $\omega_6$ is a tuning parameter, which is used to control the saliency of significant objects, such as people and cars.


\section{Experiments}
\label{sec-experiments}
In this section, we demonstrate the superiority of OCCO in infrared-visible image fusion and downstream task promotion. Firstly, in Section \ref{Setting}, we elaborate on the dataset, evaluation metrics and implementation details. Secondly, we compare with eight SOTA methods on four classical infrared and visible datasets qualitatively and quantitatively. The analysis are given in Section \ref{fusion analysis}. 

Moreover, the segmentation and detection task is also introduced to evaluate the performance of our proposed method on downstream visual task (Section \ref{Downstrem analysis}). The necessity of different modules in our proposed OCCO is discussed in Section \ref{sec:ab}. Finally, the university of contextual contrastive loss is validated in Section \ref{sec:rfn}.

\subsection{Settings}
\label{Setting}

\subsubsection{Datasets \& Metric}
\label{Datasets & Metric}
In this paper, the proposed method (OCCO) is trained on the FMB \citep{segmif} dataset, which closely resembles real road scenes with significant thermal radiation interference, resulting in abundant high-frequency information in the infrared images. This allows OCCO to differentiate important targets from other thermal noise effectively. The FMB, MSRS \citep{seafusion}, M$^3$FD \citep{tardal}, and TNO \citep{tno2014} datasets are utilized to assess the proposed method, with the numbers of image pairs for them being 280, 361, 300, and 21, respectively. 

The effectiveness of the proposed method will be validated from two aspects: pixel-level image quality and the amount of semantic information. The actual semantic validation method focus on the downstream task (Semantic Segmantation and Object Detection). Pixel-level image quality consists of four metrics: Entropy (EN) \citep{EN}, Spatial Frequency (SF) \citep{sf}, Average Gradient (AG) \citep{ag}, and Correlation Coefficient (CC) \citep{cc}. 

AG characterizes the texture details of the fused image by measuring the gradient information of the fused image. Similarly, SF reveals the details and texture information of the fused image by measuring the gradient distribution of the fused image. EN quantifies the amount of information contained in the fused image based on information theory. CC is an indicator of the linear correlation between the fused image and the source image. Higher values of these metrics indicate that the fused image contains more information and details. 

To validate the effectiveness of our method across different types of downstream tasks, we first conduct semantic segmentation on the MSRS dataset. The MSRS dataset features clear infrared targets but small image sizes, which presents a challenge for fusion methods to highlight salient objects. We perform testing after retraining for 150 epochs. Specifically, we employ the DeepLabV3+ \citep{deeplab} method on the MSRS dataset, initialize it with the same pre-trained weights intended for downstream tasks. Furthermore, to comprehensively evaluate the performance of our method, we extend our validation to object detection tasks on two multimodal datasets, MSRS and M$^3$FD. Conversely, the M$^3$FD dataset has high image clarity yet complex infrared information, demanding a strong information filtering capability. We train for 40 epochs with the same pre-trained weights on these two datasets and obtain object detection results on YOLOV5 \citep{yolov5}. The combined results from both semantic segmentation and object detection underscore the robustness and versatility of our method in processing and understanding multimodal visual information.

\subsubsection{Implementation Details}
\label{dtails}
During the training process, we randomly cropped 256$\times$256 image patches from input pairs. To improve the distinction between positive and negative samples during contrastive learning, we only select patches that contain salient objects. Ground-DINO and SAM both employ publicly available pre-trained weights for direct information extraction, operating in a zero-shot manner. Mask is pre-generated independently of the fusion network and, therefore, does not consume any resources during training or testing stage.

The Adam optimizer with a learning rate of 1e-4 is used and the batch size is set to 12. In this paper, the ratio of anchor samples to positive samples is 1, taken from the first portion of the batch, and the ratio of negative samples is 4, taken from all samples in the batch. Hyperparameters $\omega_1$, $\omega_2$, $\omega_3$, $\omega_4$, $\omega_5$ and $\omega_{6}$ are set 0.5, 0.5, 0.5, 10, 1 and 10, respectively. The proposed method is implemented based on PyTorch, and all experiments are conducted on the NVIDIA GeForce RTX 4090 GPU with 24GB memory.


\subsection{Fusion Performance Analysis}
\label{fusion analysis}
For fair comparison, there are eight state-of-the-art methods included, they are DenseFuse \citep{li2018densefuse}, SeAFusion \citep{seafusion}, LRRNet \citep{lrrnet}, CDDFuse \citep{cdd}, CoCoNet \citep{coconet}, SemLA \citep{semla},  and M2Fusion \citep{m2fusion}.

Fused images were generated using public code and pre-trained weights for each algorithm. Post-processing techniques were applied to convert grayscale images produced by DenseFuse, LRRNet and CDDFuse into color images for fair comparisons.

\subsubsection{Qualitative Results}

\begin{figure}[!ht]
	\centering
	\includegraphics[width=1\linewidth]{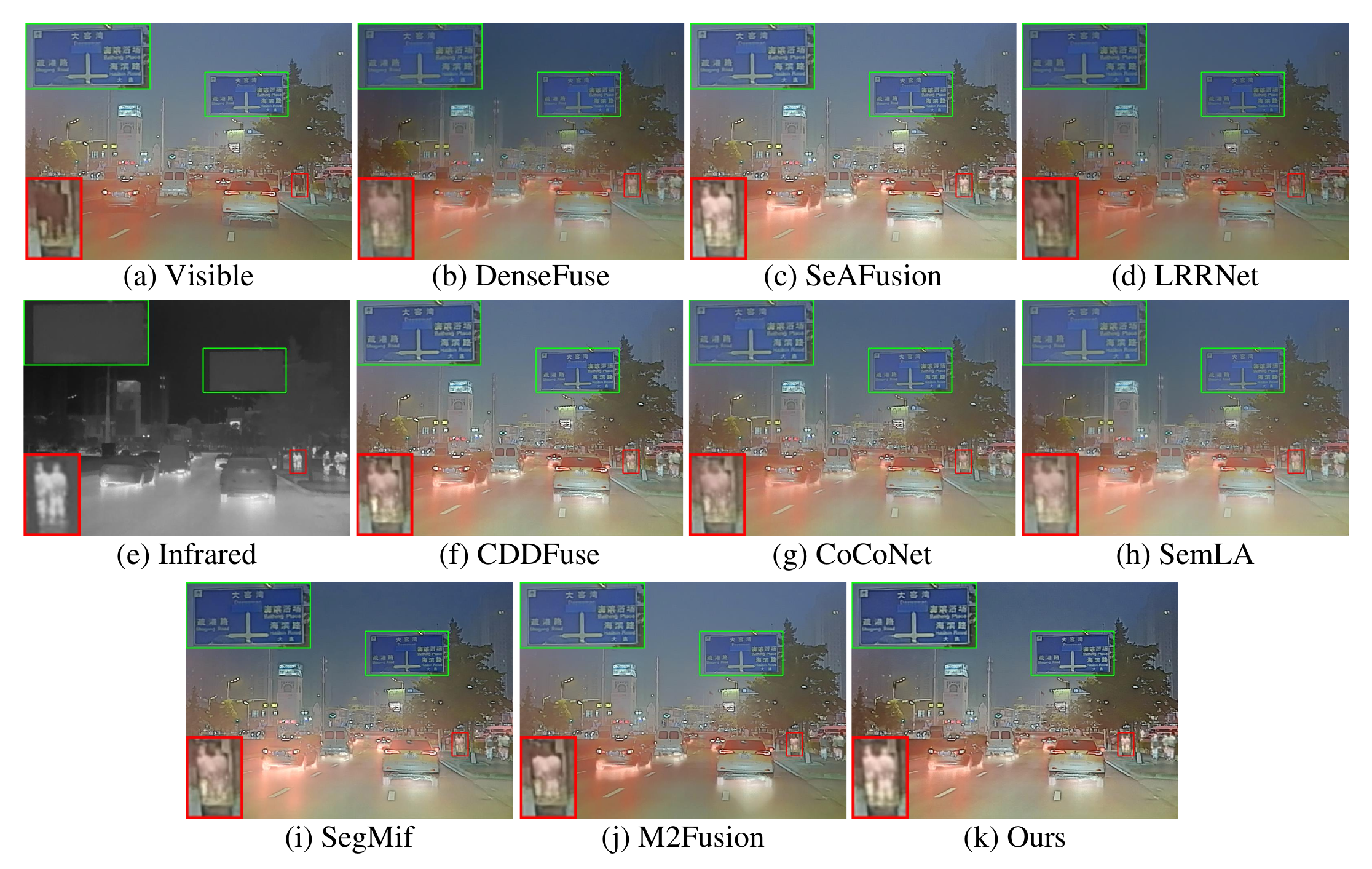}
	\caption{OCCO and eight state-of-the-art algorithms fused image of 507 image pair in the FMB dataset. Red boxes emphasize close-up pedestrian, while green boxes highlight scene details, emphasizing the differences compared to other methods. From (a) to (k) are the fusion resultes of visible image, DenseFuse, SeAFusion, LRRNet, infrared image, CDDFuse, CoCoNet, SemLA, SegMif, M2Fusion and our method.}
	\label{fig:fmb}
\end{figure}

(1) FMB dataset closely relates to real road scenes, with images of moderate size allowing detailed learning of road scene information. Visible images offer rich details, while thermal radiation in infrared images provide abundant samples for effective target differentiation. A pair of testset samples are chosen for qualitative analysis. 

In Fig.\ref{fig:fmb}, the pedestrian in the infrared image, highlighted by the red box, is partially obstructed by tree branches, resulting in incomplete shape. DenseFuse and LRRNet lose a lot of brightness information in the overall image. The other six methods all perceive high-intensity radiation on the human body, but are disturbed by scene information, leading to the dispersion of leg information. CoCoNet, SemLA, SegMif , and CDDFuse all show very blurry leg information. SeAFusion and M2Fusion maintain relatively good overall person information. However, M2Fusion surprisingly loses the head information of pedestrians. 

Finally, compared to SeAFusion, our fusion result is clear and leg information is not blurry at all. Due to some haze interference in the visible scene, the street sign marked with a green box in the visible information has some noise. Except for SegMif, none of the methods optimized this interference. Furthermore, compared to SegMif, our boundary information is more intuitive, and the street sign is clear and easy to read. In terms of salient target and scene information, our method outperforms the others.

\begin{figure}[ht]
	\centering
\includegraphics[width=0.95\linewidth]{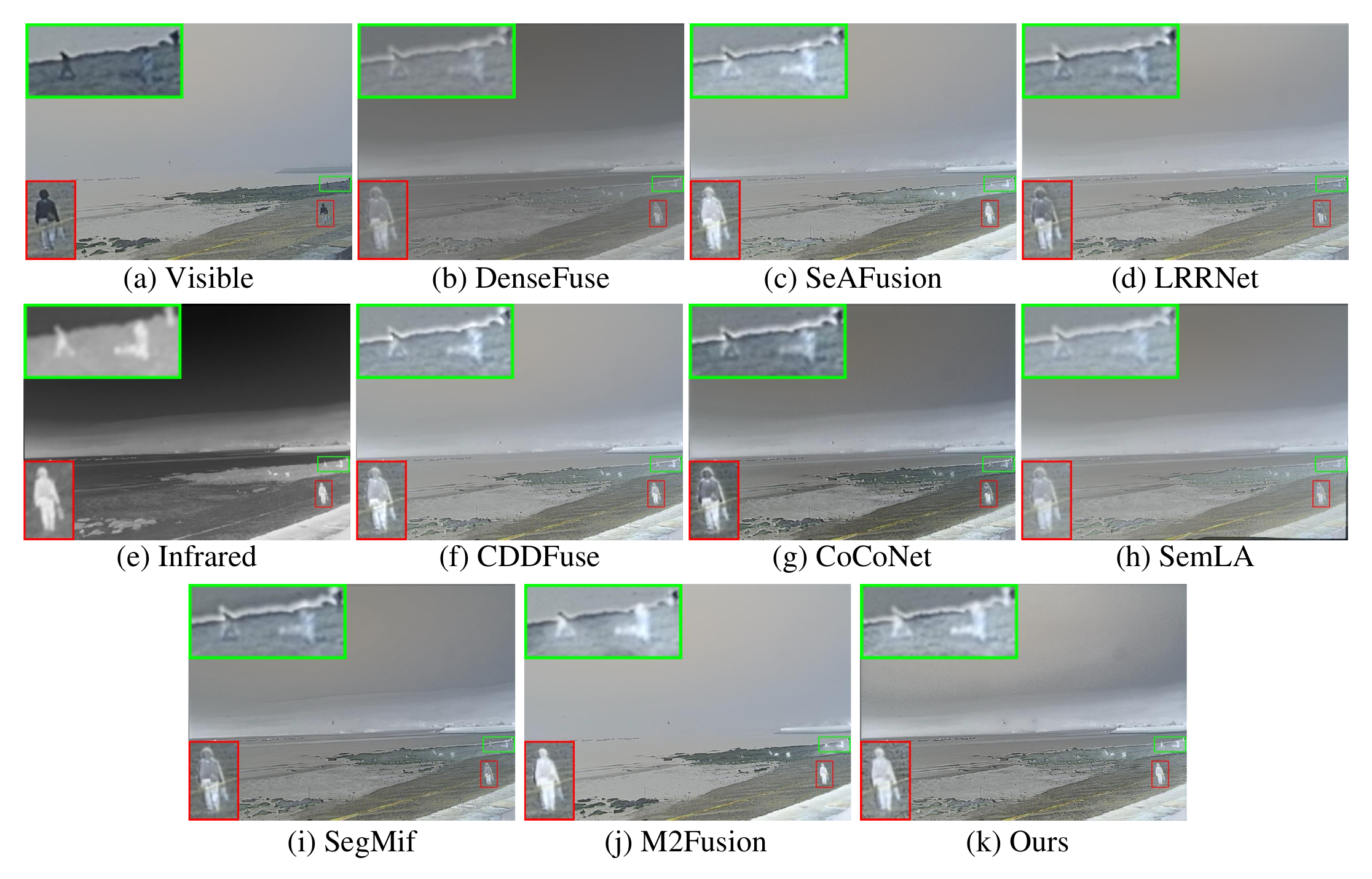}
	\caption{OCCO and eight state-of-the-art algorithms fused image of 1911 image pair in the M$^3$FD dataset. Red boxes emphasize close-up individual, while green boxes highlight people in the distance, underlining the distinctions compared to other methods. From (a) to (k) are the fusion resultes of visible image, DenseFuse, SeAFusion, LRRNet, infrared image, CDDFuse, CoCoNet, SemLA, SegMif, M2Fusion and our method.}
	\label{fig:m3fd}
\end{figure}

(2) M$^3$FD dataset, known for its high resolution, allows for the detailed observation of complex scenes. As illustrated in Fig.\ref{fig:m3fd}, an individual carrying a backpack, highlighted within the red box, may display the presence of two boundaries. This phenomenon is attributed to the interaction between the black clothing of the individual and the infrared data, leading to potential misinterpretations of the subject's outline. 
In this context, methods such as SeAFusion, M2Fusion, CDDFuse, and our proposed method demonstrate a superior ability to maintain the consistency of boundaries. In contrast, several other techniques exhibit significant seam issues, compromising the clarity of the images.

Conversely, in the grouping of individuals engaged in play, as marked by the green box, approaches such as DenseFuse, LRRNet and SegMif are noted for their tendency to merge the subjects with the background environment, resulting in faint and indistinct human silhouettes. Although SeAFusion and M2Fusion perform admirably in various aspects, it is crucial to address their limitations in preserving background information. Notably, M2Fusion completely fails to capture the delineation of the horizon where the sea meets the sky, an omission that raises concerns regarding the fidelity and realism of the produced imagery. This loss of contextual boundaries is not only unreasonable but also undermines the effectiveness of the method in accurately rendering complex scenes. 

\begin{figure}[ht]
	\centering
\includegraphics[width=0.95\linewidth]{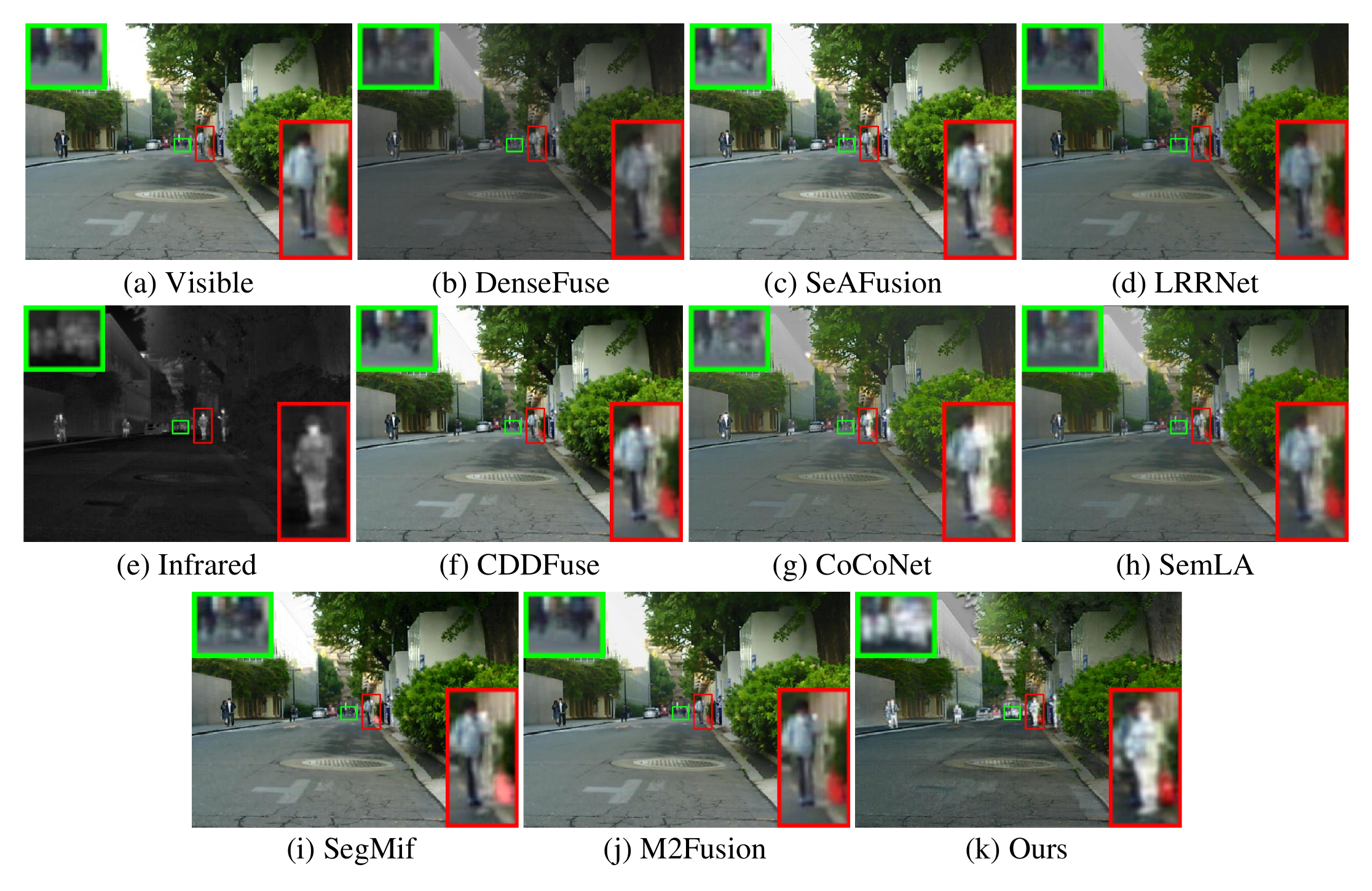}
	\caption{OCCO and eight state-of-the-art algorithms fused image of 327 image pair in the MSRS dataset. Red boxes emphasize close-up individual, while green boxes highlight pedestrians in the distance, underlining the distinctions compared to other methods. From (a) to (k) are the fusion resultes of visible image, DenseFuse, SeAFusion, LRRNet, infrared image, CDDFuse, CoCoNet, SemLA, SegMif, M2Fusion and our method.}
	\label{fig:day}
\end{figure}

(3) MSRS dataset is divided into nighttime and daytime, each showing different performance effects. 

\textbf{Daytime}, visible images exhibit rich details and abundant scene information, while infrared images highlight significant targets. However, due to device characteristics, positional discrepancies can arise for the same target across modalities. Conflicting modality information often leads to ghosting effects at individual edges in fused images, which is unfavorable for downstream tasks. As seen in Fig.\ref{fig:day}, the highlighted close-up individual in the red box suffers misalignment between modalities, with only our method and CoCoNet enhancing target infrared information, aiding humanoid shape identification. However, the training strategy of CoCoNet is not flawless and it fails to preserve boundary information completely, resulting in highly blurred fusion images, while our method can retain the complete information of the individual. With targets having low infrared brightness, other methods struggle, even SemLA, known for registration tasks, failing to align modalities perceptively. The distant group of people identified by the green box is barely visible in visible images and faintly marked in infrared. None of the other methods spot this group directly, whereas our method highlights these individuals in fusion results. At the same time, the background still maintains the information in the visible, which is in sharp contrast with the crowd. 

\begin{figure}[ht]
	\centering
\includegraphics[width=0.95\linewidth]{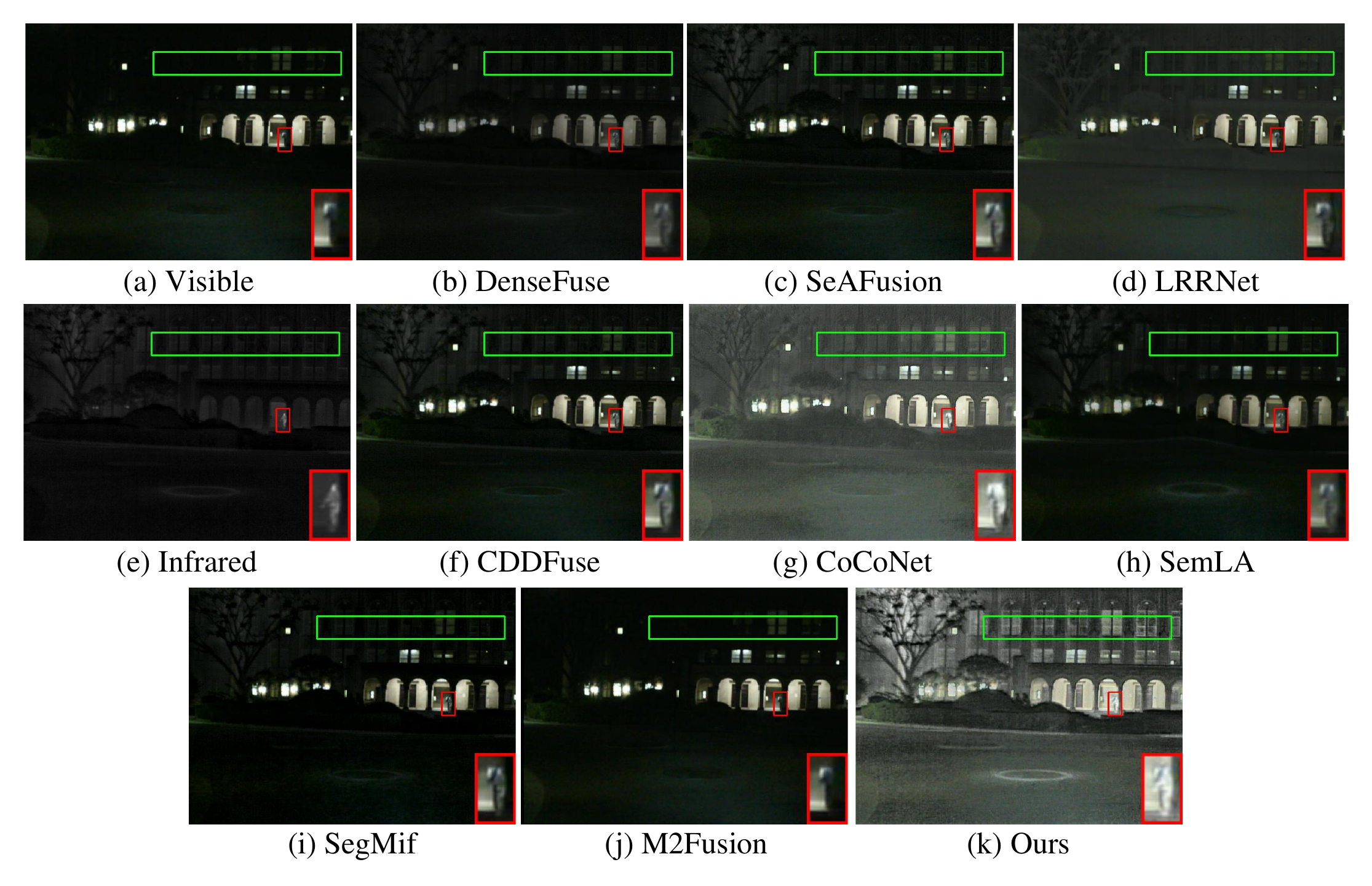}
	\caption{OCCO and eight state-of-the-art algorithms fused image of 842 image pair in the MSRS dataset. Red boxes emphasize close-up individual, while green boxes highlight sence details in the background, underlining the distinctions compared to other methods. From (a) to (k) are the fusion resultes of visible image, DenseFuse, SeAFusion, LRRNet, infrared image, CDDFuse, CoCoNet, SemLA, SegMif, M2Fusion and our method.}
	\label{fig:night}
\end{figure}

\textbf{Nighttime}, the information in the visible images is significantly weakened, and infrared information needs to be more prominent. For Fig.\ref{fig:night}, there is a person in the green box in the infrared image, similarly to the Fig.\ref{fig:day}, only CoCoNet and our method accentuate the brightness of the individual, with CoCoNet losing the human shape. M2Fusion only retains the visible information and loses all the infrared information. The other methods also favor the visible information and fail to highlight the infrared information. Our method not only highlights the person while preserving the shape of the person, only our method can see the part of the human hand. Regarding scene information, except for CoCoNet, all other methods lean towards the dark nighttime scenes depicted in visible images. The markers in the green box are challenging to discern clearly in other algorithms, whereas we can distinctly observe the presence of this marker on each pillar. 

\begin{figure}[ht]
	\centering
\includegraphics[width=0.95\linewidth]{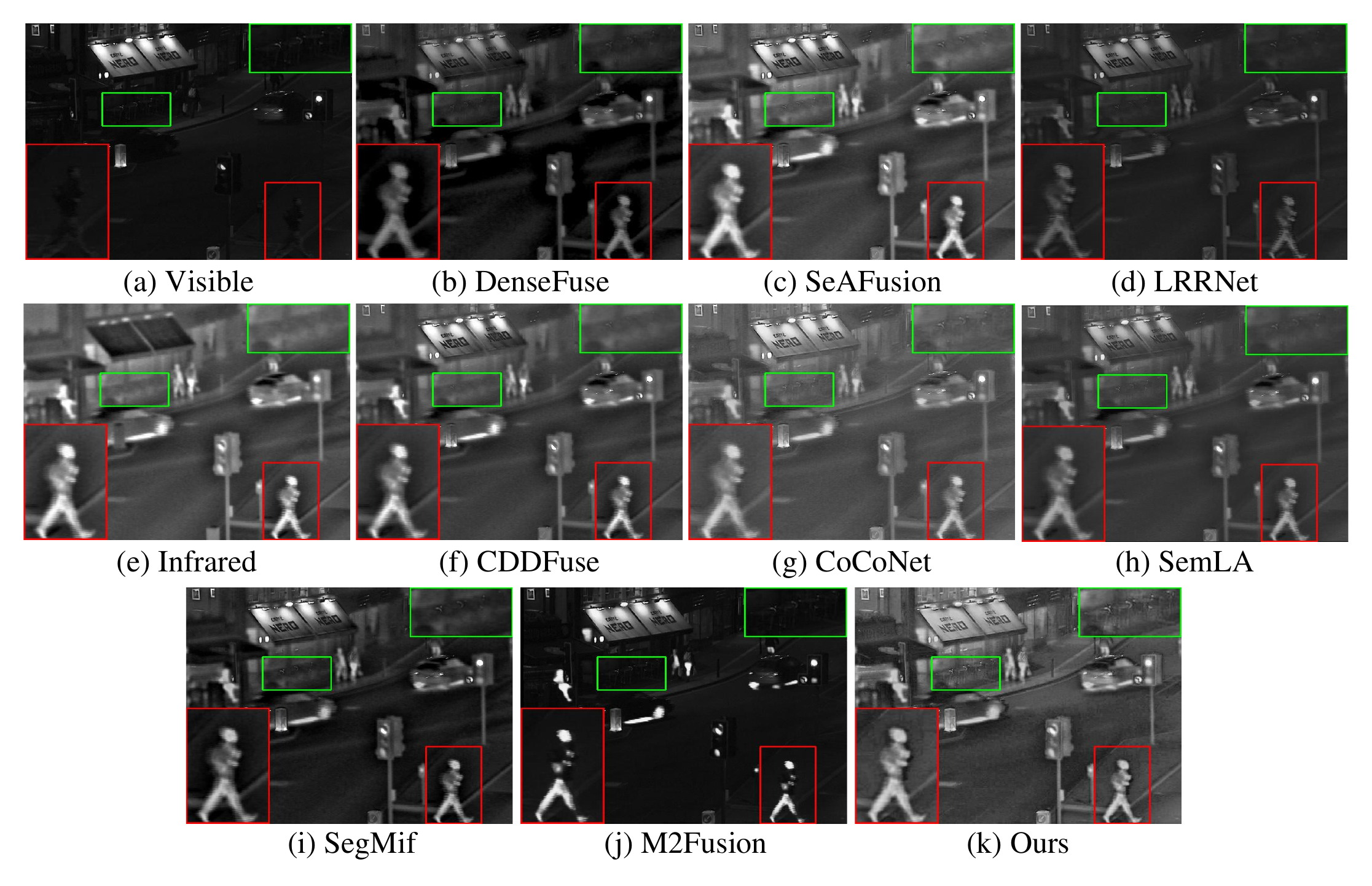}
	\caption{OCCO and eight state-of-the-art algorithms fused image of street sence in the TNO dataset. Red boxes emphasize close-up individual, while green boxes highlight sence details in the street, underlining the distinctions compared to other methods. From (a) to (k) are the fusion resultes of visible image, DenseFuse, SeAFusion, LRRNet, infrared image, CDDFuse, CoCoNet, SemLA, SegMif, M2Fusion and our method.}
	\label{fig:tno}
\end{figure}

(4) TNO dataset is characterized by simple scenes and prominently visible salient objects, allowing for a straightforward evaluation of fusion quality based on the resulting images. However, the data presented in Fig.\ref{fig:tno} exhibit a considerable degree of blurriness in the infrared information. 

Notably, our method stands out by effectively suppressing the saw-tooth noise that appears at the edges of individuals highlighted by the red box, as well as mitigating the black pseudo-shadows that emanate from those edges. Among the other approaches, CoCoNet demonstrates a commendable ability to reduce the appearance of black pseudo-shadows, although it fails to fully eliminate the disruptive saw-tooth noise. In contrast, M2Fusion exhibits a rather coarse approach to infrared information extraction, capturing only the most prominent highlights while losing critical edge details entirely.

When we shift our focus to the details observable in the visible spectrum, the chairs depicted within the store, marked by the green box, display distinct outlines. The outputs from methods such as DenseFuse, LRRNet, CDDFuse, SegMif, and M2Fusion merely capture the reflections on the armrests, failing to present a complete representation of the chairs. In contrast, CoCoNet, SeAFusion, and SemLA manage to convey a blurred impression of the chair shapes. Remarkably, only our method provides a clear depiction of the entire chair, with all four corners distinctly visible. It is particularly noteworthy that our resulting images exhibit the least amount of blurriness, highlighting the superior clarity achieved through our approach. 

Through the comparative analysis of the above four datasets with eight state-of-the-art algorithms in different scenes, OCCO can perceive salient objects and enhance their presence, while optimizing the scene information to the greatest extent.

\subsubsection{Quantitative Results} 
We compared eight state-of-the-art methods on four datasets, using four metrics (EN, AG, SF, CC), the results of FMB and M$^3$FD datasets are shown in Table \ref{tab:fmb}, the metric of MSRS and TNO datasets are shown in Table \ref{tab:msrs}. Our method consistently achieves the highest EN scores across all four datasets, indicating that our fusion images are rich in detail and complex textures, which contribute to a more uniform distribution of pixel values. 

\begin{table}[!ht]
    \caption{Quantitative results of FMB and M$^3$FD datasets. The  \textbf{bold} ,\textcolor{red}{red}and \textit{\underline{italic}} represent the best, second-best and third-best values, respectively.}
    \label{tab:fmb}
\tabcolsep=0.175cm
\begin{tabular}{
>{\columncolor[HTML]{FFFFFF}}c 
>{\columncolor[HTML]{FFFFFF}}c 
>{\columncolor[HTML]{FFFFFF}}c 
>{\columncolor[HTML]{FFFFFF}}c 
>{\columncolor[HTML]{FFFFFF}}c 
>{\columncolor[HTML]{FFFFFF}}c 
>{\columncolor[HTML]{FFFFFF}}c 
>{\columncolor[HTML]{FFFFFF}}c 
>{\columncolor[HTML]{FFFFFF}}c 
>{\columncolor[HTML]{FFFFFF}}c 
>{\columncolor[HTML]{FFFFFF}}c }
\hline
\multicolumn{1}{l}{\cellcolor[HTML]{FFFFFF}} & \multicolumn{1}{l}{\cellcolor[HTML]{FFFFFF}} & \multicolumn{4}{l}{\cellcolor[HTML]{FFFFFF}FMB}                                                                                                                & \multicolumn{1}{l}{\cellcolor[HTML]{FFFFFF}} & \multicolumn{4}{l}{\cellcolor[HTML]{FFFFFF}M$^3$FD}                                                                                                                     \\ \cline{3-6} \cline{8-11} 
{\color[HTML]{000000} Methods}               & Year                                         & EN                                    & SF                                     & AG                                    & CC                                    &                                              & EN                                          & SF                                     & AG                                    & CC                                    \\ \hline
DenseFuse                                    & 2019                                         & 6.732                                 & 8.757                                  & 2.98                                  & {\color[HTML]{FF0000} 0.652}          &                                              & 6.426                                       & 7.593                                  & 2.653                                 & {\color[HTML]{000000} \textbf{0.586}} \\
SeAFusion                                    & 2022                                         & 6.754                                 & 13.877                                 & {\ul \textit{4.275}}                  & 0.622                                 &                                              & 6.846                                       & 13.955                                 & 4.782                                 & 0.525                                 \\
LRRNet                                       & 2023                                         & 6.282                                 & 10.126                                 & 3.060                                 & {\ul \textit{0.641}}                  &                                              & 6.425                                       & 10.668                                 & 3.594                                 & 0.537                                 \\
CDDFuse                                      & 2023                                         & {\ul \textit{6.778}}                  & {\color[HTML]{FF0000} 14.577}          & {\color[HTML]{FF0000} 4.317}          & 0.631                                 &                                              & {\color[HTML]{000000} {\ul \textit{6.905}}} & {\color[HTML]{000000} \textbf{14.751}} & {\color[HTML]{FF0000} 4.870}          & 0.535                                 \\
CoCoNet                                      & 2023                                         & 6.714                                 & 12.743                                 & 3.723                                 & {\color[HTML]{000000} \textbf{0.662}} &                                              & 6.777                                       & 12.363                                 & 4.148                                 & {\color[HTML]{FF0000} 0.582}          \\
SemLA                                        & 2023                                         & 6.712                                 & 12.176                                 & 3.314                                 & 0.629                                 &                                              & 6.765                                       & 11.520                                 & 3.624                                 & 0.544                                 \\
SegMif                                       & 2023                                         & {\color[HTML]{FF0000} 6.872}          & {\ul \textit{13.926}}                  & 4.21                                  & 0.631                                 &                                              & {\color[HTML]{FF0000} 6.983}                & {\ul \textit{14.238}}                  & {\ul \textit{4.824}}                  & {\ul \textit{0.559}}                  \\
M2Fusion                                        & 2024                                         & 6.587                                 & 13.721                                 & 3.971                                 & 0.568                                 &                                              & 6.755                                       & 13.7                                   & 4.48                                  & 0.456                                 \\ \hline
ours                                         &                                              & {\color[HTML]{000000} \textbf{7.035}} & {\color[HTML]{000000} \textbf{15.095}} & {\color[HTML]{000000} \textbf{4.695}} & 0.634                                 &                                              & {\color[HTML]{000000} \textbf{6.987}}       & {\color[HTML]{FF0000} 14.249}          & {\color[HTML]{000000} \textbf{4.937}} & {\color[HTML]{330001} 0.545}          \\ \hline
\end{tabular}
\end{table}

\begin{table}[!ht]
    \caption{Quantitative results of MSRS and TNO datasets. The \textbf{bold}, \textcolor{red}{red} and \textit{\underline{italic}} represent the best, second-best and third-best values, respectively.}
    \label{tab:msrs}
\tabcolsep=0.175cm
\begin{tabular}{
>{\columncolor[HTML]{FFFFFF}}c 
>{\columncolor[HTML]{FFFFFF}}c 
>{\columncolor[HTML]{FFFFFF}}c 
>{\columncolor[HTML]{FFFFFF}}c 
>{\columncolor[HTML]{FFFFFF}}c 
>{\columncolor[HTML]{FFFFFF}}c 
>{\columncolor[HTML]{FFFFFF}}c 
>{\columncolor[HTML]{FFFFFF}}c 
>{\columncolor[HTML]{FFFFFF}}c 
>{\columncolor[HTML]{FFFFFF}}c 
>{\columncolor[HTML]{FFFFFF}}c }
\hline
\multicolumn{1}{l}{\cellcolor[HTML]{FFFFFF}} & \multicolumn{1}{l}{\cellcolor[HTML]{FFFFFF}} & \multicolumn{4}{l}{\cellcolor[HTML]{FFFFFF}MSRS}                                                                                                             & \multicolumn{1}{l}{\cellcolor[HTML]{FFFFFF}} & \multicolumn{4}{l}{\cellcolor[HTML]{FFFFFF}TNO}                                                                                                                                        \\ \cline{3-6} \cline{8-11} 
{\color[HTML]{000000} Methods}               & Year                                         & EN                                   & SF                                     & AG                                    & CC                                   &                                              & EN                                          & SF                                           & AG                                          & CC                                          \\ \hline
DenseFuse                                    & 2019                                         & 5.937                                & 6.026                                  & 2.058                                 & {\color[HTML]{000000} \textbf{0.66}} &                                              & 6.755                                       & 8.338                                        & 3.324                                       & {\color[HTML]{000000} \textbf{0.496}}       \\
SeAFusion                                     & 2022                                         & 6.65                                 & {\ul \textit{11.106}}                  & 3.697                                 & 0.609                                &                                              & {\color[HTML]{FF0000} 7.011}                & {\color[HTML]{FF0000} 11.950}                & {\color[HTML]{000000} 4.768}                & 0.457                                       \\
LRRNet                                       & 2023                                         & 6.209                                & 8.568                                  & 2.648                                 & 0.509                                &                                              & 6.815                                       & 9.393                                        & 3.641                                       & 0.434                                       \\
CDDFuse                                      & 2023                                         & {\color[HTML]{FF0000} 6.698}         & {\color[HTML]{FF0000} 11.552}          & {\color[HTML]{FF0000} 3.749}          & 0.601                                &                                              & {\color[HTML]{000000} {\ul \textit{6.936}}} & {\color[HTML]{000000} \textbf{12.092}}       & {\color[HTML]{000000} {\ul \textit{4.443}}} & 0.455                                       \\
CoCoNet                                      & 2023                                         & {\ul \textit{6.689}}                 & 10.963                                 & {\ul \textit{3.737}}                  & {\color[HTML]{000000} 0.612}         &                                              & 6.863                                       & 10.411                                       & 4.038                                       & {\color[HTML]{000000} {\ul \textit{0.461}}} \\
SemLA                                        & 2023                                         & 6.216                                & 8.311                                  & 2.686                                 & {\ul \textit{0.636}}                 &                                              & 6.541                                       & 11.360                                       & 3.400                                       & 0.404                                       \\
SegMif                                       & 2023                                         & {\color[HTML]{000000} 6.401}         & 11.005                                 & 3.61                                  & 0.608                                &                                              & {\color[HTML]{000000} 6.867}                & 11.534                                       & 4.425                                       & 0.438                                       \\
M2Fusion                                       & 2024                                         & 6.493                                & 10.874                                 & 3.185                                 & 0.542                                &                                              & 6.877                                       & 11.196                                       & 3.906                                       & 0.390                                       \\ \hline
ours                                         &                                              & {\color[HTML]{000000} \textbf{6.99}} & {\color[HTML]{000000} \textbf{11.589}} & {\color[HTML]{000000} \textbf{4.585}} & {\color[HTML]{FF0000} 0.659}         & {\color[HTML]{000000} }                      & {\color[HTML]{000000} \textbf{7.149}}       & {\color[HTML]{000000} {\ul \textit{11.714}}} & {\color[HTML]{000000} \textbf{4.937}}       & {\color[HTML]{FF0000} 0.477}                \\ \hline
\end{tabular}
\end{table}

In terms of AG, our method ranks first on MSRS, TNO, and FMB, and second on M$^3$FD, showcasing our ability to maintain clarity in nighttime scenes across these datasets, highlighting intricate details and leading to higher image clarity and richness of details. SF secures the top spot only on FMB, ranking second or third on other datasets, mainly because our method focuses on capturing the edge information of significant targets to create a distinct contrast between subjects and backgrounds, albeit with slight variations due to insufficient scene information constraints. In terms of CC, the significant targets in MSRS and TNO datasets exhibit noticeable differences from the background, enabling precise adjustment of imaging effects and proportions for various elements. However, on the large and complex road scene datasets like M$^3$FD and FMB, our model struggles to constrain scene information effectively, resulting in slightly lower performance.

\subsection{Downstream-Task Performance Analysis}
\label{Downstrem analysis}
\subsubsection{Semantic Segmantation Performance Analysis}

\begin{figure}[t]
	\centering
\includegraphics[width=1\linewidth]{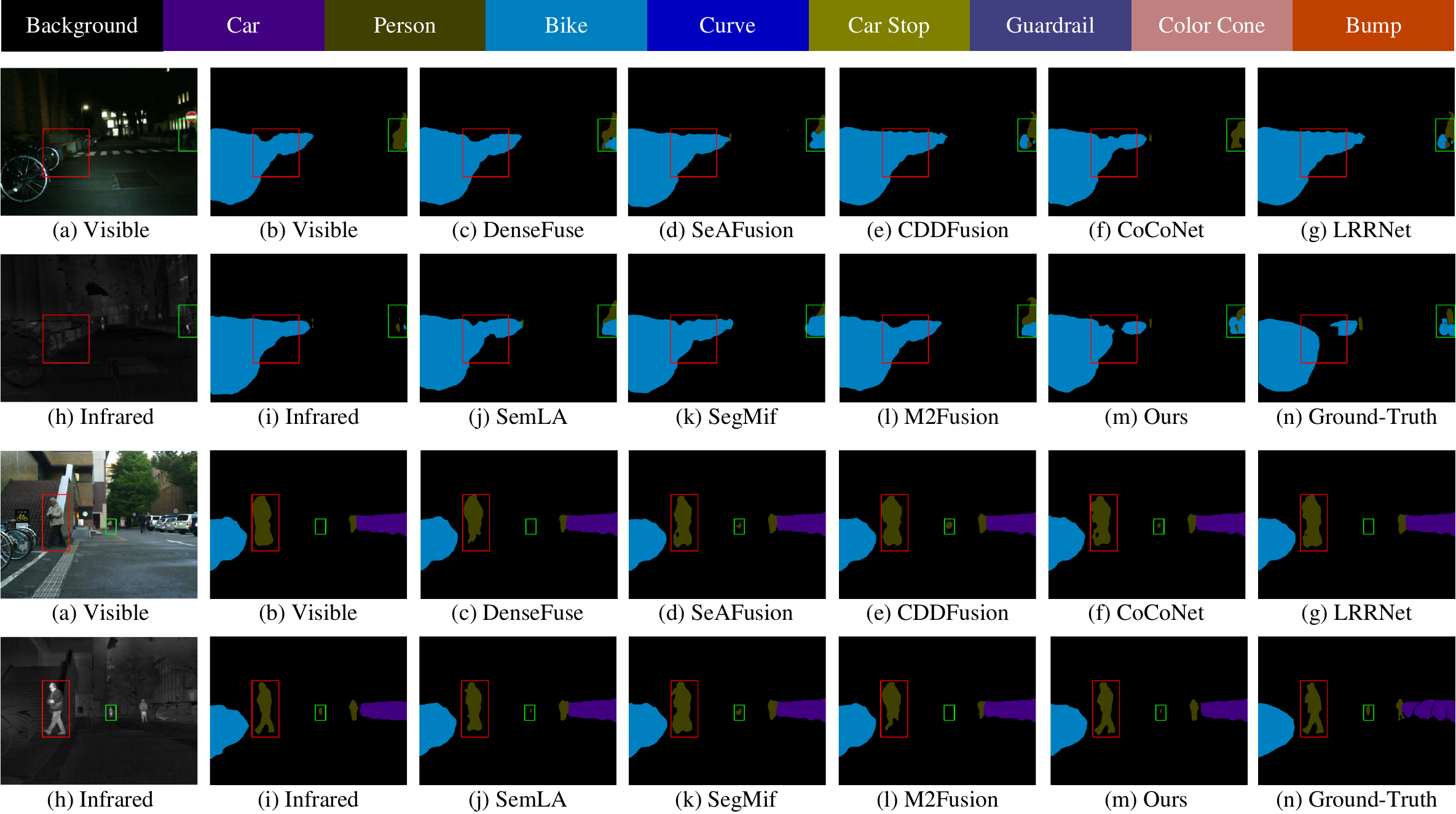}
	\caption{OCCO and eight state-of-the-art algorithms segmentation result of 1106N (first two rows) and 404D (last two row) image pair in the MSRS dataset.}
\label{fig:seg}
\end{figure}

\begin{table}[!t]
 \caption{Quantitative semantic segmentation results of MSRS. The \textbf{bold} and \textcolor{red}{red} represent the best and second-best.}
    \label{tab:seg}  
     \tabcolsep=0.0756cm
\begin{tabular}{
>{\columncolor[HTML]{FFFFFF}}c |
>{\columncolor[HTML]{FFFFFF}}c 
>{\columncolor[HTML]{FFFFFF}}c 
>{\columncolor[HTML]{FFFFFF}}c 
>{\columncolor[HTML]{FFFFFF}}c 
>{\columncolor[HTML]{FFFFFF}}c 
>{\columncolor[HTML]{FFFFFF}}c 
>{\columncolor[HTML]{FFFFFF}}c 
>{\columncolor[HTML]{FFFFFF}}c 
>{\columncolor[HTML]{FFFFFF}}c 
>{\columncolor[HTML]{FFFFFF}}c }
\hline
Method                           & Unlablled                    & Car                          & Person                                & Bike                         & Curve                        & CarStop                              & Guardrail                   & ColorCone                  & Bump                         & mIoU                          \\ \hline
Visible                          & 97.58                        & 85.51                        & 66.10                                  & 66.10                         & 35.86                        & 52.48                                 & 21.27                       & 44.13                       & 58.58                        & 58.623                        \\
{\color[HTML]{000000} Infrared}  & 97.4                         & 83.09                        & 66.88                                 & 60.09                        & {\color[HTML]{000000} 31.64} & 45.37                                 & 8.35                        & 24.93                       & 53.85                        & 52.445                          \\
{\color[HTML]{000000} DenseFuse} & 97.67                        & 85.46                        & 58.30                                  & {\color[HTML]{000000} 66.41} & 36.83                        & {\color[HTML]{000000} 57.54}          & {\color[HTML]{FF0000} 52.80} & 45.37                       & 59.81                        & 62.243                        \\
{\color[HTML]{000000} SeAFusion} & 97.83              & {\color[HTML]{FF0000} 86.91} & 65.54                                 & 65.64                        & {\color[HTML]{000000} 39.27} & {\color[HTML]{FF0000} 57.66}          & 43.65                       & 44.62                        & 58.53                         & 62.178                        \\
{\color[HTML]{000000} LRRNet}    & {\color[HTML]{000000} 97.78} & 86.35                        & 63.07                                 & 63.95                        & \textbf{42.51}               & 56.51                                 & 32.07                       & \textbf{46.35}                       & 57.23                        & 60.647                        \\
{\color[HTML]{000000} CDDFuse}   & {\color[HTML]{FF0000} 97.85} & 86.55                        & 65.05                                 & 66.28                        & 40.46                        & 55.53                                 & 40.25                       &{\color[HTML]{FF0000} 45.46}              & {\color[HTML]{FF0000} 62.59} & 62.224                        \\
{\color[HTML]{000000} CoCoNet}   & 97.81                        & {\color[HTML]{000000} 86.19} & 64.57                                 & {\color[HTML]{FF0000} 67.35} & 37.43                        & 55.45                                 & 47.42                       & 44.98                       & \textbf{63.41}               & {\color[HTML]{FF0000} 62.734} \\
{\color[HTML]{000000} SemLA}     & 97.71                        & 86.03                        & 63.15                                 & 66.52                        & 36.01                        & 54.45                                 & 34.10                        & 43.81                        & 59.89                        & 60.184                        \\
{\color[HTML]{000000} SegMif}    & 97.78                        & {\color[HTML]{000000} 86.30}  & {\color[HTML]{FF0000} 66.90}           & {\color[HTML]{000000} 66.01} & 38.58                        & 56.45                                 & 38.27                       & 44.29                       & 58.74                        & 61.48                         \\
{\color[HTML]{000000} M2Fusion}  & {\color[HTML]{000000} 97.69} & {\color[HTML]{000000} 85.76} & {\color[HTML]{000000} 58.33}          & {\color[HTML]{000000} 67.07} & {\color[HTML]{000000} 37.66} & {\color[HTML]{000000} \textbf{58.69}} & \textbf{53.58}              & 44.79                       & 60.04                        & 62.623                        \\ \hline
{\color[HTML]{000000} Ours}      & \textbf{97.90}                & \textbf{87.02}               & {\color[HTML]{000000} \textbf{67.59}} & \textbf{67.98}               & {\color[HTML]{FF0000} 42.32} & 55.27                                 & 52.22                       & 45.43 & 57.84                         & \textbf{63.72}                \\ \hline
\end{tabular}
\end{table}
We evaluated the semantic segmentation performance using the MSRS dataset. DeepLabV3+ serves as a representative baseline, illustrating the semantic information extraction capability inherent in various fusion algorithms. The segmentation results are shown in Fig.\ref{fig:seg}. 

A significant challenge arises in nighttime scenes due to the absence of visible information. Specifically, within the red-boxed regions indicating bicycles, infrared modality also failed to provide a salient representation, thereby posing a challenge to achieving cohesive boundary delineation. However, our method uniquely succeeded in discerning the interstitial gaps between bicycles. In contrast, other approaches erroneously classified the obstructions situated between adjacent rows of bicycles as bicycles themselves.  Furthermore, a circular road sign positioned directly above a cyclist’s head on the right side of the image presented a distraction that adversely affected most methodologies. Only CoCoNet, SegMif, and our method effectively constrained the segmentation target within the accurate boundaries.  Nevertheless, it is noteworthy that while achieving precise localization, these methods concurrently suffered from a reduced capacity to detect the bicycles themselves. Of particular interest is the presence of a distant pedestrian located centrally within the image. Our method, along with SeAFusion, CoCoNet, and SemLA, successfully detected this subtle and diminutive object. 

In daytime scenarios, the dark trousers of pedestrians in visible image exhibit a high degree of visual similarity to the background, complicating differentiation; this issue is further compounded for pedestrians at a distance. Addressing this necessitates fusion algorithms to meticulously preserve detailed boundaries derived from infrared information.  Consistent with this requirement, our method commendably preserved clear boundaries, with the refined segmentation of the leg regions being particularly pronounced.  Consequently, the distant pedestrian was accurately segmented, an outcome not realized by alternative methods.

Table \ref{tab:seg} corroborates that our method achieves the best semantic segmentation performance for common salient object categories, including pedestrians, vehicles, and roads.  Although it exhibits slightly weaker performance in categories like ``curve" and ``color cone", it still secures the second and third positions. Therefore, overall, our method maintains superior performance.

\subsubsection{Object Detection Performance Analysis}
\label{detection analysis}
\begin{figure}[!t]
	\centering
\includegraphics[width=0.75\linewidth]{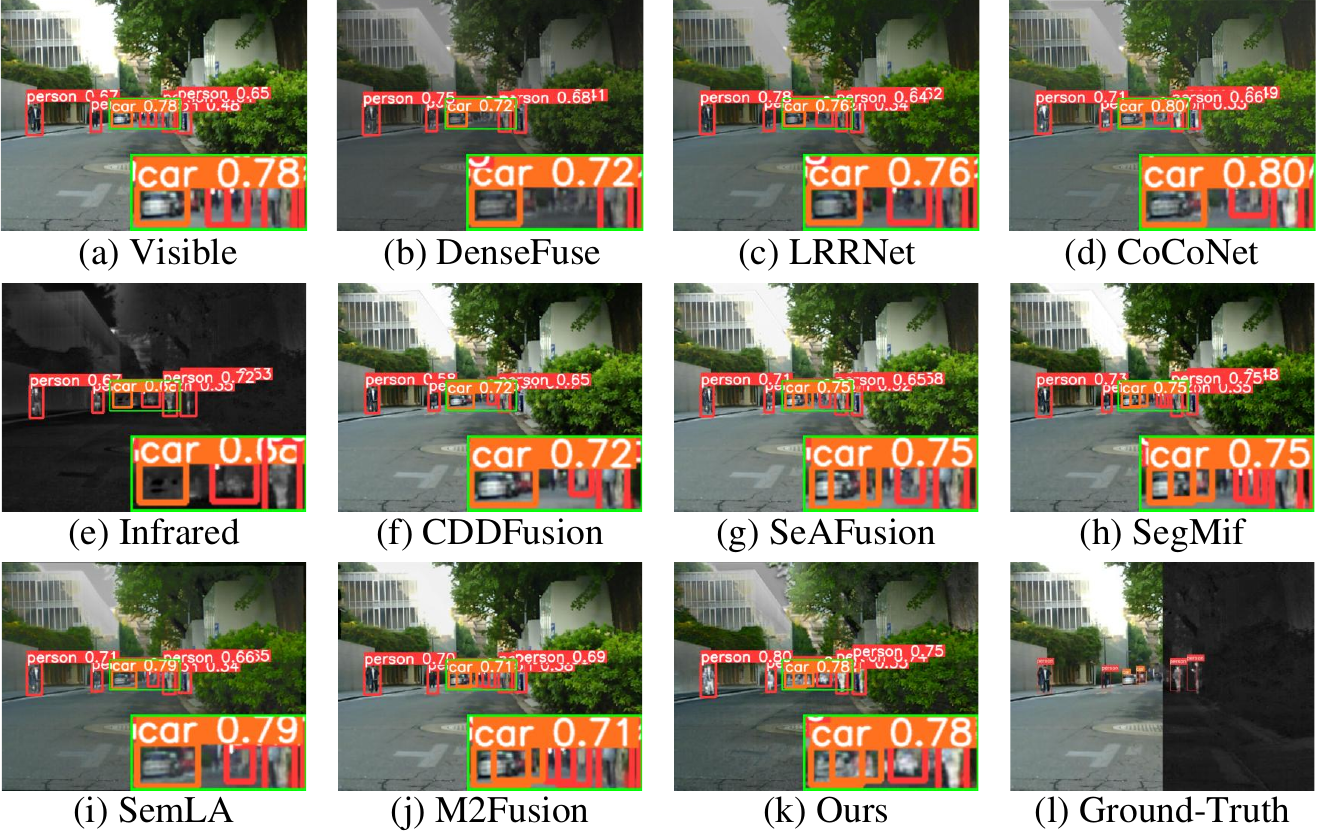}
	\caption{OCCO and eight state-of-the-art algorithms detection result of 327D image pair in the MSRS dataset.}
\label{fig:detection}
\end{figure}

\begin{figure}[t]
	\centering
\includegraphics[width=0.75\linewidth]{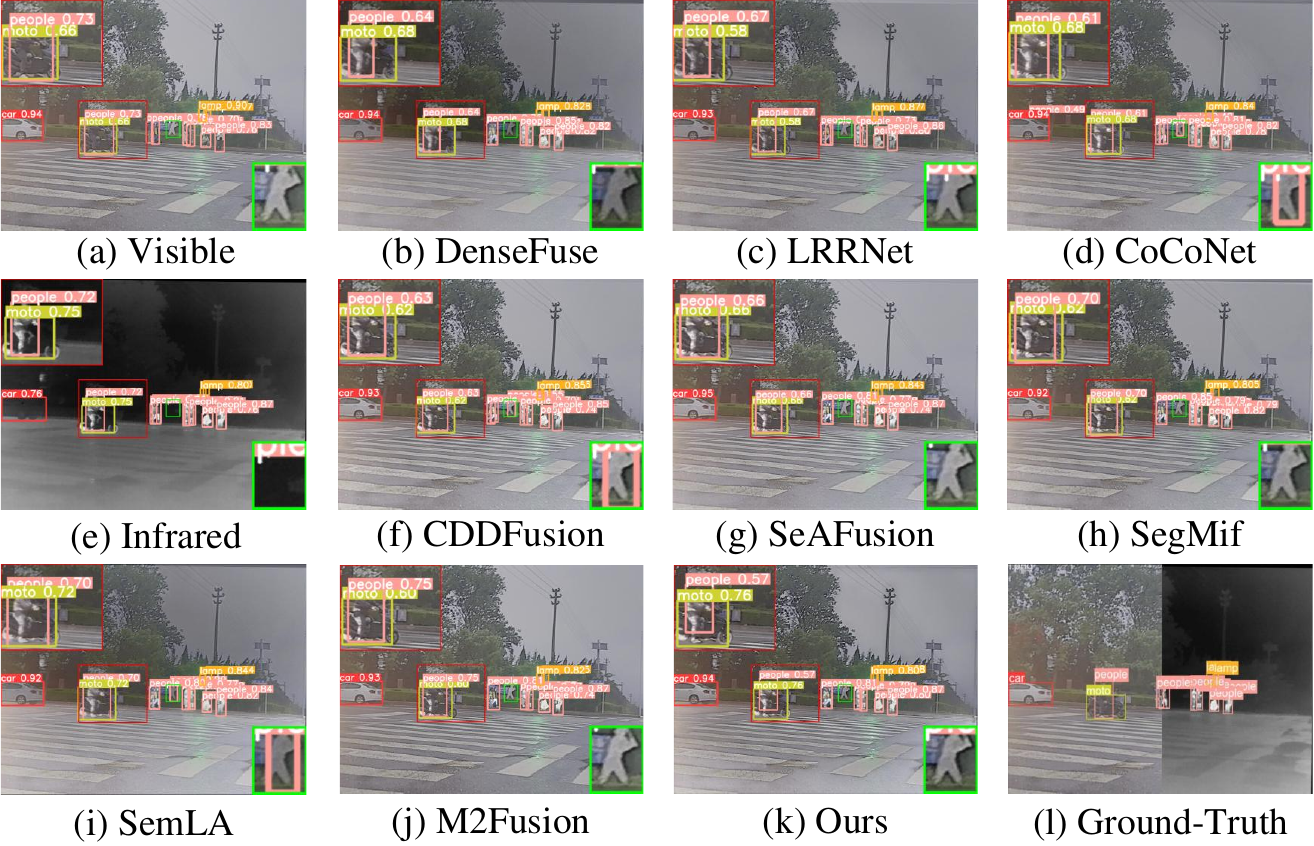}
	\caption{OCCO and eight state-of-the-art algorithms detection results of 198 image pair in the M$^3$FD dataset. }
	\label{fig:m3fdde}
\end{figure}
Even though labels from MSRS dataset do not annotate pedestrians far out as shown in Fig.\ref{fig:detection}, after training, except for DenseFuse, all models can detect these pedestrians. The most significant difference is reflected in the detection of vehicles, where only our approach along with SeAFusion \citep{seafusion}, SegMif \citep{segmif}, and M2Fusion \citep{m2fusion} annotated two cars. Clearly, our confidence is the highest, surpassing the original image and all other methods. The MSRS dataset also includes bicycle classification, but due to low label quality, all methods only achieve around 40\% in terms of mAP, so they are not included in the statistics. As OCCO learns specifically for people and vehicles, we achieved first place in the quantitative detection results for people and car in Table \ref{tab:detection}. 

\begin{table}[t]
 \caption{Quantitative detection results of MSRS. The \textbf{bold} and \textcolor{red}{red} represent the best and second-best.}
    \label{tab:detection}  
\begin{tabular}{c|cc|cc|cc}
\hline
                                                         & \multicolumn{2}{c|}{person}                                                   & \multicolumn{2}{c|}{car}                                                      & \multicolumn{2}{c}{all}                                                       \\ \cline{2-7} 
\multirow{-2}{*}{Method}                                 & mAp                                   & mAP@.5:.95                            & mAp                                   & mAP@.5:.95                            & mAp                                   & mAP@.5:.95                            \\ \hline
\cellcolor[HTML]{FFFFFF}{\color[HTML]{000000} Visible}   & 0.511                                 & 0.211                                 & 0.773                                 & 0.431                                 & 0.642                                 & 0.321                                 \\
\cellcolor[HTML]{FFFFFF}{\color[HTML]{000000} Infrared}  & 0.803                                 & 0.41                                  & 0.776                                 & 0.415                                 & 0.790                                 & 0.413                                 \\
\cellcolor[HTML]{FFFFFF}{\color[HTML]{000000} DenseFuse} & 0.720                                 & 0.349                                 & 0.712                                 & 0.382                                 & 0.716                                 & 0.366                                 \\
\cellcolor[HTML]{FFFFFF}{\color[HTML]{000000} SeAFusion } & {\color[HTML]{FF0000} 0.796}          & 0.410                                 & 0.758                                 & 0.428                                 & 0.777                                 & 0.419                                 \\
\cellcolor[HTML]{FFFFFF}{\color[HTML]{000000} LRRNet}    & 0.767                                 & 0.379                                 & 0.764                                 & 0.429                                 & 0.766                                 & 0.404                                 \\
\cellcolor[HTML]{FFFFFF}{\color[HTML]{000000} CDDFuse}   & 0.806                                 & {\color[HTML]{FF0000} 0.416}          & 0.759                                 & 0.423                                 & 0.783                                 & 0.420                                 \\
\cellcolor[HTML]{FFFFFF}{\color[HTML]{000000} CoCoNet}   & 0.775                                 & 0.381                                 & 0.745                                 & 0.413                                 & 0.760                                 & 0.397                                 \\
\cellcolor[HTML]{FFFFFF}{\color[HTML]{000000} SemLA}     & 0.706                                 & 0.337                                 & 0.771                                 & {\color[HTML]{FF0000} 0.447}          & 0.739                                 & 0.392                                 \\
\cellcolor[HTML]{FFFFFF}{\color[HTML]{000000} SegMif}    & 0.799                                 & 0.410                                 & 0.775                                 & 0.446                                 & {\color[HTML]{FF0000} 0.787}          & {\color[HTML]{FF0000} 0.428}          \\
\cellcolor[HTML]{FFFFFF}{\color[HTML]{000000} M2Fusion}  & 0.632                                 & 0.280                                 & {\color[HTML]{FF0000} 0.775}          & 0.438                                 & 0.704                                 & 0.359                                 \\ \hline
\cellcolor[HTML]{FFFFFF}{\color[HTML]{000000} ours}      & {\color[HTML]{000000} \textbf{0.811}} & {\color[HTML]{000000} \textbf{0.437}} & {\color[HTML]{000000} \textbf{0.794}} & {\color[HTML]{000000} \textbf{0.472}} & {\color[HTML]{000000} \textbf{0.803}} & {\color[HTML]{000000} \textbf{0.455}} \\ \hline
\end{tabular}
\end{table}
In the M$^3$FD dataset, the infrared information is too noisy, which is not conducive to extracting portraits. Table \ref{tab:m3de} shows that OCCO ranks second only to the infrared mode itself but performs well in the category of vehicles, which is infrared modality lacks. However, due to a lack of attention to other scene information, the detection performance for ``lamp'' is poor. Overall, we have achieved the best performance. As shown in Fig.\ref{fig:m3fdde}, CDDFuse, CoCoNet, and SeAFusion erroneously recognized the background boards of human shapes as people. Apart from riders, our confidence in detecting ``person'' exceeded 79\%, significantly higher than other methods, and we also performed well in recognizing motorcycles.

\begin{table}[!t]
\caption{Quantitative detection results of M$^3$FD. The \textbf{bold} and \textcolor{red}{red} represent the best and second-best.}
\label{tab:m3de}
\tabcolsep=0.15cm
\begin{tabular}{
>{\columncolor[HTML]{FFFFFF}}c |
>{\columncolor[HTML]{FFFFFF}}c 
>{\columncolor[HTML]{FFFFFF}}c |
>{\columncolor[HTML]{FFFFFF}}c 
>{\columncolor[HTML]{FFFFFF}}c |
>{\columncolor[HTML]{FFFFFF}}c 
>{\columncolor[HTML]{FFFFFF}}c |
>{\columncolor[HTML]{FFFFFF}}c 
>{\columncolor[HTML]{FFFFFF}}c }
\hline
\cellcolor[HTML]{FFFFFF}                         & \multicolumn{2}{c|}{\cellcolor[HTML]{FFFFFF}car}                              & \multicolumn{2}{c|}{\cellcolor[HTML]{FFFFFF}people}                          & \multicolumn{2}{c|}{\cellcolor[HTML]{FFFFFF}truck}                            & \multicolumn{2}{c}{\cellcolor[HTML]{FFFFFF}lamp}            \\ \cline{2-9} 
\multirow{-2}{*}{\cellcolor[HTML]{FFFFFF}Method} & mAp                                   & mAp.5:.95                             & mAp                                   & mAp.5:.95                            & mAp                                   & mAp.5:.95                             & mAp                          & mAp.5:.95                    \\ \hline
{\color[HTML]{000000} Visible}                   & 0.844                                 & 0.543                                 & 0.564                                 & 0.249                                & {\color[HTML]{FF0000} 0.652}          & 0.367                                 & \textbf{0.6}                 & \textbf{0.255}               \\
{\color[HTML]{000000} Infrared}                  & 0.819                                 & 0.515                                 & \textbf{0.705}                        & \textbf{0.358}                       & \textbf{0.655}                        & {\color[HTML]{FF0000} 0.38}           & 0.399                        & 0.156                        \\
{\color[HTML]{000000} DenseFuse}                 & \textbf{0.851}                        & {\color[HTML]{FF0000} 0.552}          & 0.665                                 & 0.326                                & 0.645                                 & 0.377                                 & 0.584                        & \textbf{0.255}               \\
{\color[HTML]{000000} SeAFusion}                 & {\color[HTML]{FF0000} 0.846}          & 0.543                                 & 0.669                                 & 0.326                                & 0.624                                 & 0.358                                 & 0.544                        & 0.221                        \\
{\color[HTML]{000000} LRRNet}                    & 0.844                                 & 0.545                                 & 0.649                                 & 0.312                                & 0.645                                 & 0.371                                 & {\color[HTML]{FF0000} 0.595} & {\color[HTML]{FF0000} 0.251} \\
{\color[HTML]{000000} CDDFuse}                   & 0.842                                 &  0.544         & 0.654                                 & 0.323                                & 0.629                                 & 0.362                                 & 0.542                        & 0.236                        \\
{\color[HTML]{000000} CoCoNet}                   & 0.843                                 & 0.544                                 & 0.652                                 & 0.315                                & 0.647                                 & 0.368                                 & 0.576                        & 0.243                        \\
{\color[HTML]{000000} SemLA}                     & 0.843                                 & 0.539                                 & 0.594                                 &  0.268        & 0.63                                  & 0.36                                  & 0.576                        & 0.239                        \\
{\color[HTML]{000000} SegMif}                    & {\color[HTML]{FF0000} 0.846}          & {\color[HTML]{FF0000} 0.552}          & 0.658                                 & 0.322                                & 0.649          & 0.361          & 0.591                        & 0.246                        \\
{\color[HTML]{000000} M2Fusion}                  & 0.833                                 & 0.537                                 &  0.633          & 0.306                                & 0.624                                 & 0.362                                 & 0.545                        & 0.225                        \\ \hline
{\color[HTML]{000000} ours}                      & 0.844 & {\color[HTML]{000000} \textbf{0.553}} & {\color[HTML]{FF0000} 0.681} & {\color[HTML]{FF0000} 0.34} & 0.642 & {\color[HTML]{000000} \textbf{0.386}} & 0.586                        & 0.250                         \\ \hline
\hline

\hline
\cellcolor[HTML]{FFFFFF}                         & \multicolumn{2}{c|}{\cellcolor[HTML]{FFFFFF}moto}           & \multicolumn{2}{c|}{\cellcolor[HTML]{FFFFFF}bus}                             & \multicolumn{2}{c|}{\cellcolor[HTML]{FFFFFF}all}                                \\ \cline{2-7} 
\multirow{-2}{*}{\cellcolor[HTML]{FFFFFF}Method} & mAp                          & mAp.5:.95                    & mAp                                   & mAp.5:.95                            & mAp                                    & mAp.5:.95                             \\ \cline{1-7} 
{\color[HTML]{000000} Visible}                   & 0.489                        & \textbf{0.254}               & 0.77                                  & 0.556                                & {\color[HTML]{000000} 0.6532}          & 0.3707                                \\
{\color[HTML]{000000} Infrared}                  & 0.466                        & 0.227                        & 0.782                                 & {\color[HTML]{000000} 0.577}         & 0.6377                                 & {\color[HTML]{000000} 0.3688}         \\
{\color[HTML]{000000} DenseFuse}                 & \textbf{0.492}               & {\color[HTML]{000000} 0.238} & 0.769                                 & 0.561                                & {\color[HTML]{FF0000} 0.6677}          & {\color[HTML]{FF0000} 0.3848}         \\
{\color[HTML]{000000} SeAFusion}                 & {\color[HTML]{000000} 0.462} & 0.25                         & 0.752                                 & 0.555                                & 0.6495                                 & 0.3755                                \\
{\color[HTML]{000000} LRRNet}                    & 0.444                        & 0.225                        & 0.776                                 & 0.552                                & 0.6588                                 & 0.376                                 \\
{\color[HTML]{000000} CDDFuse}                   & 0.488                        & {\color[HTML]{000000} 0.247} & 0.765                                 & 0.553                                & 0.6533                                 & 0.3775                                \\
{\color[HTML]{000000} CoCoNet}                   & 0.481                        & 0.242                        & 0.762                                 & 0.552                                & 0.6602                                 & 0.3773                                \\
{\color[HTML]{000000} SemLA}                     & 0.483                        & {\color[HTML]{FF0000} 0.251} & 0.769                                 & {\color[HTML]{000000} 0.546}         & 0.6492                                 & 0.3672                                \\
{\color[HTML]{000000} SegMif}                    & {\color[HTML]{000000} 0.461}  & {\color[HTML]{000000} 0.224}  & {\color[HTML]{FF0000} 0.784}          & {\color[HTML]{FF0000} 0.579}         & {\color[HTML]{000000} 0.6647}          & {\color[HTML]{000000} 0.3800}           \\
{\color[HTML]{000000} M2Fusion}                  & 0.453                        & 0.244                        & {\color[HTML]{000000} 0.784}          & 0.565                                & 0.6453                                 & 0.3732                                \\ \cline{1-7} 
{\color[HTML]{000000} ours}                      & {\color[HTML]{FF0000} 0.491}  & {\color[HTML]{000000} 0.250}  & {\color[HTML]{000000} \textbf{0.804}} & {\color[HTML]{000000} \textbf{0.58}} & {\color[HTML]{000000} \textbf{0.6745}} & {\color[HTML]{000000} \textbf{0.393}} \\ \cline{1-7} 
\end{tabular}
\end{table}

\subsection{Ablation Studies}
\label{sec:ab}
\subsubsection{Influence of Contextual Contrastive Loss}
\label{sec:cl}

\begin{figure}[t]
	\centering
\includegraphics[width=1\linewidth]{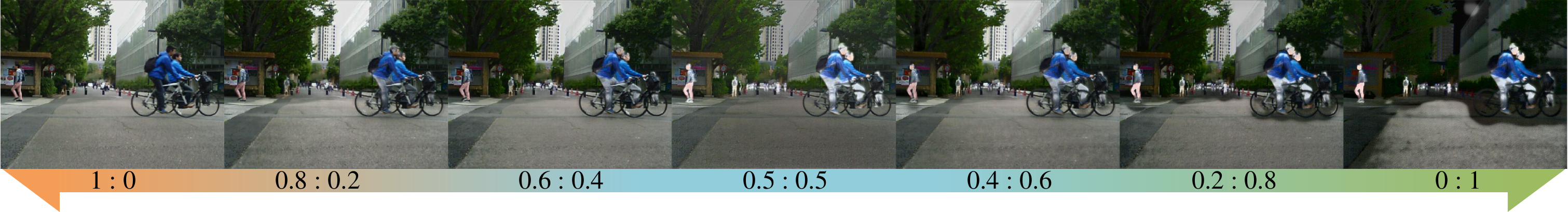}
	\caption{The impact of varying weight allocations, where ``1:0'' denotes the configuration with $\omega_{1}$ (visible information) set to 1 and $ \omega_{2}$ (infrared information) set to 0.  In this paper, ``0.5:0.5'' represents our proposed method.}
\label{fig:balance}
\end{figure}

\textbf{Bias in Shared Information} To validate the bias in shared information within the fusion results, we first conduct experiments by adjusting the settings of \( \omega_{1} \) and \( \omega_{2} \) in Equation.\ref{balance}. Since the allocation of \( \omega_{1} \) and \( \omega_{2} \) is a matter of proportionality, their sum is constrained to 1. Specifically, the configuration ``1:0" denotes that $\omega_1$ is set to 1 and $\omega_2$ to 0, resulting in the shared information being entirely derived from visible modality, with no contribution from infrared modality. The experimental results for these settings are evaluated in Figure 3. Following this, we systematically adjust the weights by decreasing $\omega_1$ from 1 to 0 in increments of 0.2, while correspondingly increasing $\omega_2$ from 0 to 1, to investigate the effects across intermediate configurations. The experimental results for these settings are evaluated in Figure \ref{fig:balance}.

\begin{table}[!t]
\caption{Quantitative results of bias of shared information. The \textbf{bold} and \textcolor{red}{red} represent the best and second-best.}
\label{tab:baalance}
\begin{tabular}{c|cccc}
\hline
\rowcolor[HTML]{FFFFFF} 
\begin{tabular}[c]{@{}c@{}}$\omega_{1} :\omega_{2}$\end{tabular} & EN                                    & SF                                     & AG                                    & CC                                    \\ \hline
\rowcolor[HTML]{FFFFFF} 
1:0                                                                                                  & {\color[HTML]{000000} 7.003}          & {\color[HTML]{000000} \textbf{16.000}} & {\color[HTML]{000000} \textbf{5.242}} & {\color[HTML]{000000} 0.501}          \\
\rowcolor[HTML]{FFFFFF} 
{\color[HTML]{000000} 0.8:0.2}                                                                       & {\color[HTML]{000000} 7.157}          & {\color[HTML]{000000} 13.905}          & {\color[HTML]{000000} 4.593}          & {\color[HTML]{FF0000} 0.516}          \\
\rowcolor[HTML]{FFFFFF} 
0.6:0.4                                                                                              & 7.090                                 & 14.535                                 & 4.659                                 & 0.497                                 \\
\rowcolor[HTML]{FFFFFF} 
{\color[HTML]{000000} \textbf{0.5:0.5}}                                                                       & {\color[HTML]{000000} \textbf{7.035}} & {\color[HTML]{FF0000} 15.095}          & {\color[HTML]{FF0000} 4.695}          & {\color[HTML]{000000} \textbf{0.634}} \\
\rowcolor[HTML]{FFFFFF} 
{\color[HTML]{000000} 0.4:0.6}                                                                       & {\color[HTML]{000000} 7.040}          & {\color[HTML]{000000} 13.163}          & {\color[HTML]{000000} 4.193}          & {\color[HTML]{000000} 0.512}          \\
\rowcolor[HTML]{FFFFFF} 
{\color[HTML]{000000} 0.2:0.8}                                                                       & {\color[HTML]{000000} 7.134}          & {\color[HTML]{000000} 5.019}           & {\color[HTML]{000000} 1.771}          & {\color[HTML]{000000} 0.380}          \\
\rowcolor[HTML]{FFFFFF} 
{\color[HTML]{000000} 0:1}                                                                           & {\color[HTML]{FF0000} 7.245}          & {\color[HTML]{000000} 12.671}          & {\color[HTML]{000000} 4.085}          & {\color[HTML]{000000} 0.462}          \\ \hline
\end{tabular}
\end{table}

``0:1" setting demonstrably exhibits significant artifacts and informational conflicts. This phenomenon is attributed to our background design, which exhibits a strong affinity with visible information, thereby creating a stark contrast with infrared information. In contrast, the ``1:0" setting effectively outputs an image closely resembling the original visible image, thus maintaining harmony with the background context. It can be observed when the proportion of visible information ($\omega_1$) is predominant, the overall harmony is relatively high with fewer artifacts; however, the saliency of the target remains insufficient. Conversely, when the proportion of infrared information ($\omega_2$) dominates, the artifacts surrounding the salient target become more pronounced, although the target itself is more distinctly highlighted. Neither of these outcomes represents an ideal result.

Ultimately, we observe that the balanced ``0.5 : 0.5" configuration effectively accommodates the differences between information modalities, resolving conflicts with background information and enabling a greater focus on information cohesion. Comparing the target information across different configurations, the ``0.5 : 0.5" setting yields the cleanest and most coherent information in the average state. Quantitative results in the Table \ref{tab:baalance} reveal that the extreme ``1 : 0" and ``0 : 1" configurations achieve optimal EN, SF, and AG metrics. However, these peak values are misleading and not practically meaningful, thus excluded from further consideration. Among the remaining configurations, the balanced ``0.5 : 0.5" setting effectively achieves a harmonious state between targets and background, minimizing artifacts, which contributes to its superior performance across these three metrics. Based on this analysis, we set \( \omega_{1}=0.5 \) and \( \omega_{2}=0.5 \) in the subsequent experiments.
\\ \hspace*{\fill} \\

\noindent\textbf{Proportion of Contextual Contrastive Loss}
Our method jointly constrains the fusion results at both the pixel and semantic levels, with $\omega_{6}$ adjusted as a parameter to balance these two constraints. The contextual contrastive loss is designed to enhance the integrity of target information and improve the semantic richness of target representations. We select four magnitudes of weight for validation.

\begin{figure}[t]
	\centering
\includegraphics[width=0.5\linewidth]{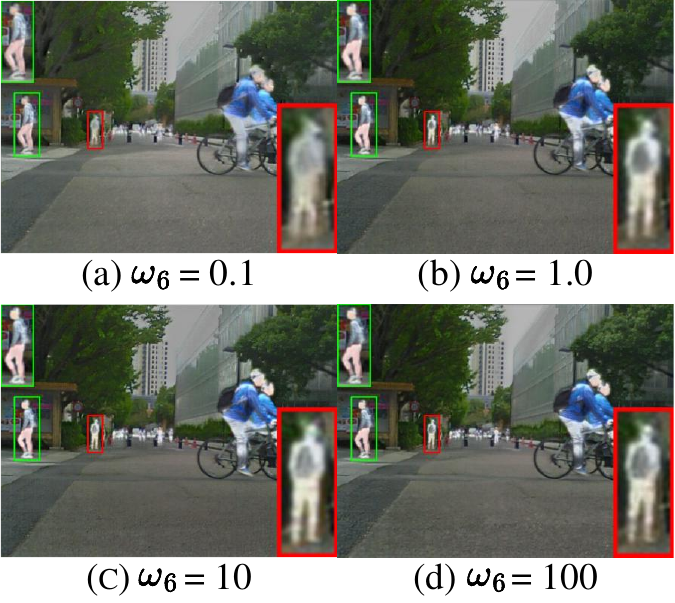}
	\caption{The impact of proportion of contextual contrastive loss. $\omega_6=10$ is our proposed method.}
\label{fig:clear}
\end{figure}

Fusion results are visualized in Fig.\ref{fig:clear}. Consistent with our design rationale, when the weight of contextual contrastive loss is relatively small, the overall semantic coherence capability is not robust. This is particularly evident in the backpacker, highlighted within the red box. With the weight set to 0.1, the black backpack merges indistinctly with the person, and boundaries are lost. When the weight value is increased to 1.0, initial improvements are observed; the distinction between the backpack and the person becomes more apparent. However, edge information around the person still exhibits loss.

\begin{table}[t]
\caption{Quantitative results of proportion of contextual contrastive loss. The \textbf{bold} and \textcolor{red}{red} represent the best and second-best.}
\label{tab:clear}
\begin{tabular}{c|cccc}
\hline
\rowcolor[HTML]{FFFFFF} 
$\omega_{6}$             & EN                           & SF                                                    & AG                           & CC                           \\ \hline
\rowcolor[HTML]{FFFFFF} 
0.1                      & {\color[HTML]{FF0000} 6.949} & {\color[HTML]{000000} 12.329}                         & {\color[HTML]{000000} 4.018} & {\color[HTML]{000000} 0.619} \\
\rowcolor[HTML]{FFFFFF} 
{\color[HTML]{000000} 1} & {\color[HTML]{000000} 6.909} & \cellcolor[HTML]{FFFFFF}{\color[HTML]{FF0000} 13.580} & {\color[HTML]{000000} 4.147} & {\color[HTML]{000000} 0.622} \\
\rowcolor[HTML]{FFFFFF} 
\textbf{10}                       & \textbf{7.035}               & {\color[HTML]{000000} \textbf{15.095}}                & {\color[HTML]{FF0000} 4.695} & \textbf{0.634}               \\
\rowcolor[HTML]{FFFFFF} 
100                      & 6.870                        & 13.314                                                & \textbf{4.723}               & {\color[HTML]{FF0000} 0.623} \\ \hline
\end{tabular}
\end{table}
Further increasing the weight to a magnitude of 10 yields clearer edges and a complete backpack – a significantly improved outcome.  When the weight is further increased to 100, we observe highly coherent person information; even the outline of a black hat on the head becomes visible, and the backpack contour is at its sharpest. However, this over-coherence comes at the cost of information diffusion, such as in the leg regions, which is not a desirable performance. From a metric perspective, while AG shows a partial improvement at a weight of 100, this contradicts our initial objective of mitigating information diffusion. To address this, and considering the trade-off between AG metric improvement and information diffusion control, our method ultimately employs a contextual contrastive loss weight allocation of 10.
\\ \hspace*{\fill} \\

\begin{figure}[t]
	\centering
\includegraphics[width=0.85\linewidth]{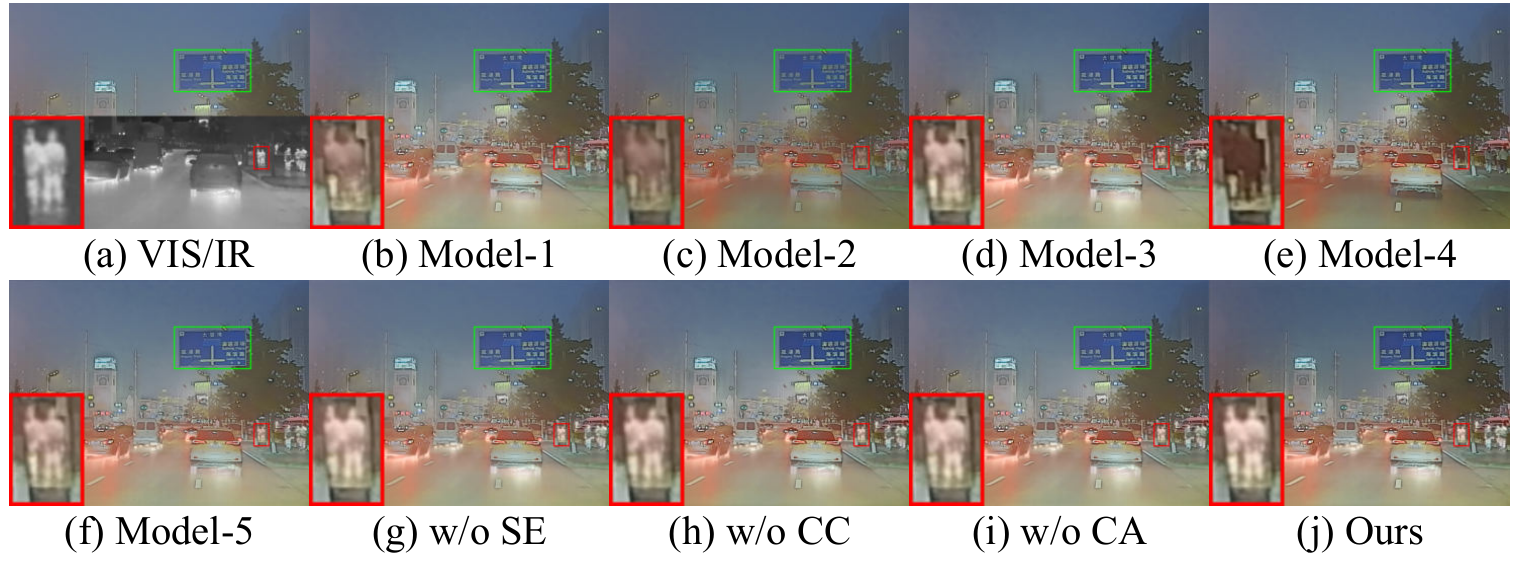}
	\caption{The fusion results of ablation experiments of 507 image pairs in the FMB dataset. (a) Details in visible and infrared images, (b) Constraining only deep features in Euclidean  space, (c) Constraining all features in contextual space, (d) Constraining both deep and shallow features in Euclidean space, (e) Directly calculating intensity loss without contrastive learning, (f) Constraining only relative distances in contextual space while disregarding absolute distances, (g) Excluding spatial enhancement modules, (h) Excluding channel enhancement modules, (i) Excluding cross-attention modules, (j) The method proposed in this paper.  }
	\label{fig:ablation}
\end{figure}
\noindent\textbf{Design of Contextual Space} Contrastive learning commonly adopts the Euclidean space as the laten feature space, utilizing VGG19 to extract features from samples. Anchor samples are chosen based on task requirements to pull features closer or push them further apart. Contextual Contrastive Learning constrains deep features only in the contextual space, thus introducing ablation $\mathtt{Model -1}$: restricting deep features solely in the Euclidean space. Not constraining shallow features is to emphasize learning deep semantic information and relax the suppression of shallow features. Subsequently, ablation $\mathtt{Model -2}$ is set up to compare the five-layer features in the contextual space, and ablation $\mathtt{Model- 3}$ compares the five-layer features in the Euclidean space. Given the pre-generation of numerous salient object masks, ablation $\mathtt{Model- 4}$ is designed to eliminate the form of comparative learning by applying the masks to intensity losses, the loss function as follows:
\begin{equation}
L_{abl}=\omega_1\|f_{\overline{vi}}-vi_{\overline{vi}}\|_2+\omega_2\|f_{\overline{ir}}-vi_{\overline{ir}}\|_2+\|f_{\overline{bg}}-vi_{\overline{bg}} \|_2,
\label{ab}
\end{equation}

Eq.\ref{ab} replace $L_{con}$ and $L_{int}$. Calculations are made for the relative distances between samples while also constraining the absolute distances between samples, leading to the final model concerning contextual space, $\mathtt{Model- 5}$, which solely restricts relative distances without assistance from absolute distances, that means $\omega_{3}=0$.

The ablation results, as shown in the Fig.\ref{fig:ablation}, indicate that $\mathtt{Model- 1}$ and $\mathtt{Model- 3}$ demonstrate that the Euclidean space necessitates collective constraints on all features. Solely constraining deep features in the Euclidean space leads to feature dispersion, resulting in sparse leg information in the final fusion outcome. Even when comparing all features, the Euclidean distance fails to aggregate infrared information. $\mathtt{Model- 2}$ reveals a significant influence of the contextual space on relative distances; demanding closeness in relative distances even in shallow features causes the fusion result to be more blurred, with severe loss of details. 

\begin{figure}[t]
\setlength{\abovecaptionskip}{0.cm}
	\centering
\includegraphics[width=0.75\linewidth]{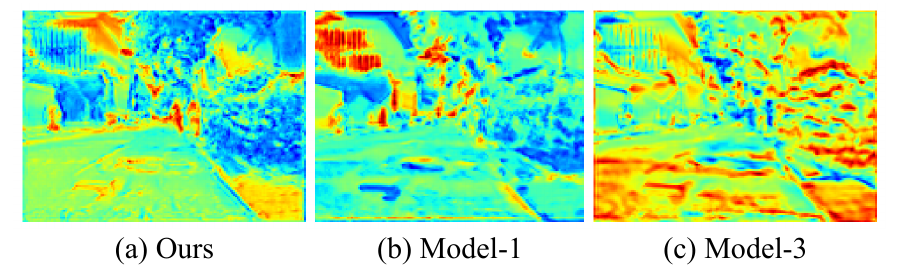}
	\caption{Feature visualization of the fusion result of 0327D image pairs in different feature Spaces in the MSRS dataset. (a) The method proposed in this paper, (b) Constraining only deep features in Euclidean feature space, (c) Constraining all features in Euclidean space.}
	\label{fig:feature}
\end{figure}

The visualization of fused features obtained by different latent feature space are shown in Fig.\ref{fig:feature}. Contextual space highlights all individuals in red, ensuring their completeness and differentiation from the background. In the Euclidean space, constraints on either only deep features ($\mathtt{Model- 1}$ ) or all features ($\mathtt{Model- 3}$ ) fail to focus attention on individuals, leading to some loss of targets. 

Back to Fig.\ref{fig:ablation}, without the form of contrastive learning, masks lose their significance, and directly applying intensity loss constraints is insufficient to encourage the model to learn bimodal information, leading to the complete loss of the infrared information, as shown in $\mathtt{Model- 4}$. In the latent feature space, relying solely on relative distances to enhance target integrity results in a loss of saliency, as demonstrated in $\mathtt{Model- 5}$. Therefore, the proposed contextual space, compared to traditional Euclidean space, can enhance target aggregation and improve saliency significantly. From Table \ref{tab:abl}, it can be observed that contrastive learning based on Euclidean distance falls slightly short in terms of entropy, resulting in some loss of correlation. Departing from the form of contrastive learning and directly using mask constraints lead to a decrease in overall metrics, as the absence of absolute distance constraints causes a significant loss of spatial frequency.


\subsubsection{Influence of Network Structure}
\label{abl-net}

\begin{table}[t]
    \caption{Quantitative results of contextual space and network , the models are explained in the Section \ref{sec:cl} and Section \ref{abl-net}. The \textbf{bold} are the best.}
    \label{tab:abl}
\begin{tabular}{ccccc}
\hline
\rowcolor[HTML]{FFFFFF} 
Model                                              & EN                                                & SF                                                 & AG                                                & CC                                                \\ \hline
\rowcolor[HTML]{FFFFFF} 
$\mathtt{Model -1}$                                           & 6.881                                              & 13.714                                             & 4.189                                             & 0.62                                              \\
\rowcolor[HTML]{FFFFFF} 
$\mathtt{Model -2}$                                            & 6.942                                             & 13.633                                             & 4.176                                             & 0.631                                             \\
\rowcolor[HTML]{FFFFFF} 
$\mathtt{Model -3}$                                            & 6.798                                             & 14.894                                             & 4.522                                             & 0.480                                             \\
\rowcolor[HTML]{FFFFFF} 
$\mathtt{Model -4}$                                           & 6.423                                             & 11.809                                             & 3.527                                             & 0.556                                             \\
\rowcolor[HTML]{FFFFFF} 
$\mathtt{Model -5}$                                            & 6.864                                             & 12.903                                             & 4.025                                             & 0.623                                             \\ \hline
\rowcolor[HTML]{FFFFFF} 
\multicolumn{1}{l}{\cellcolor[HTML]{FFFFFF}    w/o SE} & \multicolumn{1}{l}{\cellcolor[HTML]{FFFFFF}6.853}  & \multicolumn{1}{l}{\cellcolor[HTML]{FFFFFF}13.347} & \multicolumn{1}{l}{\cellcolor[HTML]{FFFFFF}3.995} & \multicolumn{1}{l}{\cellcolor[HTML]{FFFFFF}0.623} \\
\rowcolor[HTML]{FFFFFF} 
\multicolumn{1}{l}{\cellcolor[HTML]{FFFFFF}    w/o CC} & \multicolumn{1}{l}{\cellcolor[HTML]{FFFFFF}6.981}  & \multicolumn{1}{l}{\cellcolor[HTML]{FFFFFF}14.626} & \multicolumn{1}{l}{\cellcolor[HTML]{FFFFFF}4.503} & \multicolumn{1}{l}{\cellcolor[HTML]{FFFFFF}0.607} \\
\rowcolor[HTML]{FFFFFF} 
\multicolumn{1}{l}{\cellcolor[HTML]{FFFFFF}    w/o CA} & \multicolumn{1}{l}{\cellcolor[HTML]{FFFFFF}6.918} & \multicolumn{1}{l}{\cellcolor[HTML]{FFFFFF}14.064} & \multicolumn{1}{l}{\cellcolor[HTML]{FFFFFF}4.223} & \multicolumn{1}{l}{\cellcolor[HTML]{FFFFFF}0.621} \\ \hline
\rowcolor[HTML]{FFFFFF} 
ours                                               & \textbf{7.035}                                    & \textbf{15.095}                                    & \textbf{4.695}                                    & \textbf{0.634}                                    \\ \hline
\end{tabular}
\end{table}


As a core component of the Feature Interaction Fusion Network, the Feature Interaction Fusion Block (FIFB) plays a crucial role in understanding the network's structural efficacy through ablation studies. The FIFB comprises three integral components: Spatial Enhancement (SE), Cross Channel (CroC), and Cross Attention (CA). Each of these elements contributes distinctly to the overall performance of the network, underscoring their individual and collective importance in achieving high-quality image fusion.

Starting with Spatial Enhancement (SE), our ablation studies reveal that omitting this component results in pronounced artifacts appearing at the overall scene boundaries. This effect is particularly marked in the middle of the road in Fig.\ref{fig:ablation}, where utility poles are situated, which significantly detracts from the clarity of the image. The absence of SE not only compromises visual appeal but also leads to a substantial decline in performance metrics such as Spatial Fidelity (SF), Aggregate Gradation (AG), and Entropy (EN), as evidenced in Table \ref{tab:abl}. This highlights SE's crucial role in refining the spatial structure and detail of the fused images.

In addition, the CroC component is essential for maintaining the correlation between various features across different channels. The findings from our experiments indicate that removing CroC results in a significant drop in the Correlation Coefficient of the entire image. This decline illustrates the critical function of CroC in ensuring that inter-channel relationships are preserved, which is vital for enriching the quality and coherence of the final output. By integrating cross-channel information, the network effectively enhances feature interaction, leading to improved image fidelity.
\begin{figure}[t]
\setlength{\abovecaptionskip}{0.cm}
	\centering
\includegraphics[width=0.75\linewidth]{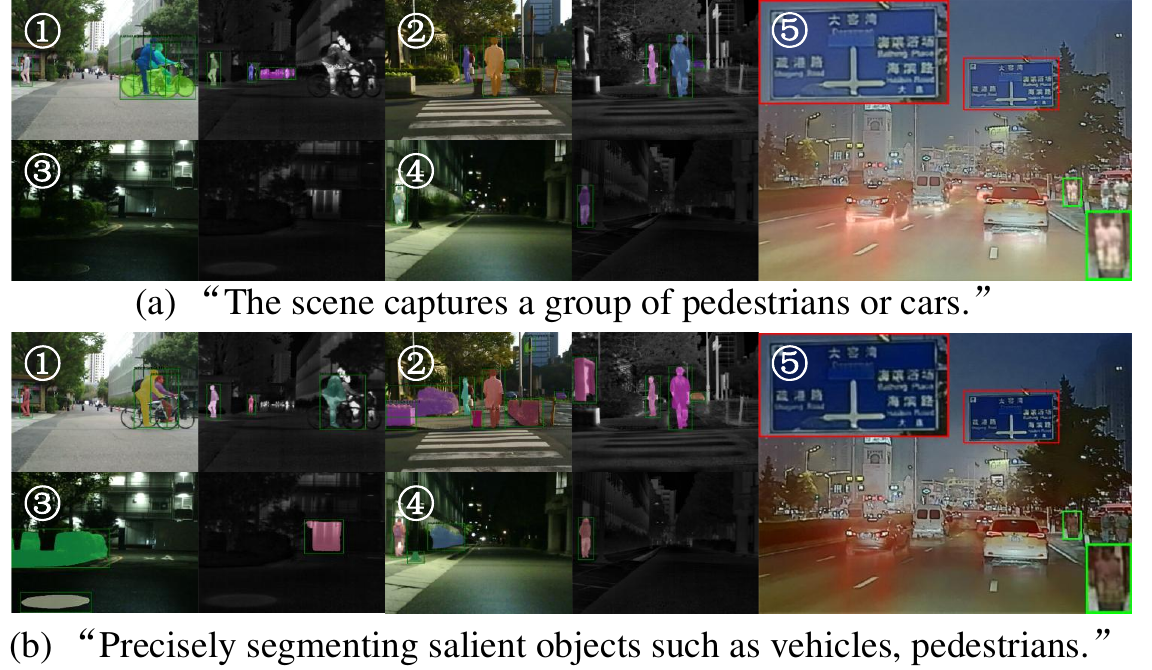}
	\caption{Impact of different text prompts on LVM information extraction; samples \ding{172}, \ding{173}, \ding{174}, and \ding{175} show information extraction results in different scenarios, while sample \ding{176} shows the fusion result trained under this text prompt.}
	\label{fig:sam_ab}
\end{figure}
Furthermore, the Cross Attention (CA) mechanism serves to further elevate the overall quality of the fused images. Our analysis with the w/o CA variant shows a noticeable degradation in image quality, emphasizing that CA is vital for enabling the model to dynamically focus on pertinent features within the dataset. This mechanism allows the network to prioritize critical elements while suppressing less relevant information, thereby optimizing the fusion process and resulting in higher-quality images.

Overall, the ablation experiments clearly demonstrate the indispensable contributions of Spatial Enhancement, Cross Channel, and Cross Attention components within the Feature Interaction Fusion Block. Each of these elements plays a vital role, as their absence significantly hampers the performance and effectiveness of the Feature Interaction Fusion Network, ultimately underscoring their importance in achieving superior image fusion results.

\subsubsection{Influence of LVM-Guide}
\label{sec:sam}

\begin{table}[t]
    \caption{Quantitative results of different text prompt. The \textbf{bold} are the best.}
    \label{tab:sam}
\begin{tabular}{c|cccc}
\hline
\rowcolor[HTML]{FFFFFF} 
Text Prompt                   & EN                                    & SF                                     & AG                                    & CC                                    \\ \hline
\rowcolor[HTML]{FFFFFF} 
All salient object                       & {\color[HTML]{000000} \textbf{7.037}}          & {\color[HTML]{000000} 14.684}          & {\color[HTML]{000000} 4.648}          & {\color[HTML]{000000} 0.498}          \\
\rowcolor[HTML]{FFFFFF} 
{\color[HTML]{000000} Only person and cars} & {\color[HTML]{000000} 7.035} & {\color[HTML]{000000} \textbf{15.095}} & {\color[HTML]{000000} \textbf{4.695}} & {\color[HTML]{000000} \textbf{0.634}} \\ \hline
\end{tabular}
\end{table}

The LVM serves as crucial semantic guidance, and the information it provides is pivotal to our approach. To demonstrate the significance of LVM, we conduct experiments employing varying text prompts. In our configuration, we utilize the text prompt: ``\textit{The scene captures a group of pedestrians or cars.}" This prompt is designed to direct the LVM's attention to pedestrians and cars, the most prevalent semantic categories in road scenes. This choice is predicated on the fact that these object classes typically exhibit well-defined boundaries in visible image. Concurrently, in thermal infrared modality, pedestrians and cars are often characterized by high-intensity thermal signatures, rendering them exceptionally amenable to reliable detection and segmentation.

Conversely, we explore an alternative configuration wherein we incorporate the keyword ``\textit{salient}" into the text prompt, thereby removing constraints on the LVM's attention. The modified prompt is: ``\textit{Precisely segmenting salient objects such as vehicles, pedestrians.}" The information extraction outcomes are visually presented in Fig.\ref{fig:sam_ab}, with \ding{172} and \ding{173} representing daytime scenarios, and \ding{174} and \ding{175} depicting nighttime scenarios. In Fig.\ref{fig:sam_ab} (b), the LVM indeed captures a greater volume of information. However, this expanded information capture is accompanied by the introduction of numerous low-quality samples. For instance, sample (a)-\ding{174}, which was originally considered a discarded or irrelevant sample, is now incorporated into the learning process under the new prompt in (b)-\ding{174}. Similarly, a substantial number of obstacles in (b)-\ding{173} are also included. These low-quality samples collectively bias the fusion results towards attending to a broader spectrum of scene information.

Referring back to the source images in Fig.\ref{fig:fmb}, we observe in the final fusion result (b)-\ding{176} that scene information is indeed maximized. This maximization effectively mitigates interference from fog present in the visible image and excessive thermal radiation noise from the ground. However, this emphasis on comprehensive scene information comes at the expense of neglecting a significant portion of salient objects, a detrimental trade-off in the context of road scene understanding. Furthermore, a comparative analysis of the quantitative metrics presented in the Table \ref{tab:sam} reveals that the implementation of the new text prompt does not yield any improvement in overall fusion quality.

\begin{figure}[t]
\setlength{\abovecaptionskip}{0.cm}
	\centering
\includegraphics[width=1\linewidth]{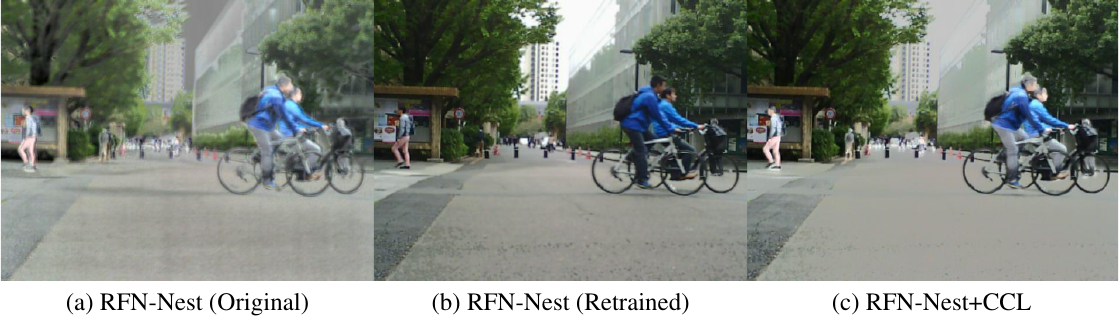}
	\caption{(a) Fusion results from RFN-Nest utilizing publicly available pre-trained weights. (b)  Fusion results from RFN-Nest re-trained on MSRS dataset. (c) Fusion results from RFN-Nest re-trained on MSRS dataset with Contextual Contrastive Loss}
	\label{fig:rfn}
\end{figure}

\subsection{Universality of Contextual Contrastive Loss}
\label{sec:rfn}
Due to RFN-Nest’s feature loss being similar to the concept of Contextual Contrastive Loss and the design of the network architecture is simple and convenient for training, we conduct ablation experiments using RFN-Nest\citep{rfnnest} as a baseline model to rigorously validate the Universality of contextual contrastive loss.

Initially, we directly employ the publicly available pre-trained weights of RFN-Nest, which are originally trained on the KAIST\citep{kaist}. Upon direct testing with these pre-trained weights, observed that the fusion results exhibited pronounced artifacts in Fig.\ref{fig:rfn}(a). Furthermore, owing to the inherent design of the original loss function as below, 

\begin{equation}
    L_{R F N}=\alpha\left(1-\operatorname{SSIM}\left(I_{f}, I_{v i}\right)\right)+L_{ {feature }} 
\end{equation}
\begin{equation}
    L_{\text {feature }}=\sum_{m=1}^{M} w_{1}(m)\left\|\Phi_{f}^{m}-\left(w_{v i} \Phi_{v i}^{m}+w_{i r} \Phi_{i r}^{m}\right)\right\|_{F}^{2}
\end{equation}

Here, $SSIM$ represents the structural similarity loss,  $I_f$ is the fusion image, $I_{vi}$ is the original visible image, $\Phi_f^m$  is the intermediate feature, and $\|\cdot\|_{F}$ is the Frobenius norm. To ensure a fairer evaluation, we re-train the RFN-Nest from scratch on the MSRS dataset. The results of this re-training are presented in Fig.\ref{fig:rfn}(b). As can be observed, the fusion outputs from the re-trained RFN-Nest model exhibit a strong resemblance to the visible source image. Building upon this re-trained baseline, we subsequently integrated Contextual Contrastive Loss and re-trained the RFN-Nest. The loss function is:
\begin{equation}
    L_{RFN+CCL}=L_{RFN}+\beta L_{CCL}.
\end{equation}
Where $\beta$ is used to balance the magnitude of the two loss functions. As depicted in Fig.\ref{fig:rfn}(c), the fusion results from RFN-Nest+CCL demonstrably show a significant enhancement in the representation of salient objects and clear boundary information.

\section{Conclusion}
\label{sec-con}
This paper introduces a novel LVM-guided image fusion method which can balance the performance of high-level visual task and the quality of fused image. Initially, a large vision model, SAM, is used to generate the semantic masks from two different modalities, decomposing the masks into modality-unique and modality-shared information based on their information components. Various semantic masks are employed to extract rich modality samples, driving the fused images to preserve semantic information through contextual contrastive learning. Samples with more semantic content are considered positive, while those with sparse semantic content are considered negative, with the fused images as anchor samples. Furthermore, a new latent feature space (contextual space) is proposed, aggregating features on a per-feature point basis, which significantly enhances feature coherence and saliency compared to the Euclidean space. 

The quantitative experiments on four datasets compared with eight state-of-the-art algorithms demonstrate that the fusion results obtained by the proposed method achieve better fusion performance. The qualitative experiments show that our fusion method can obtain more brightness and more complete boundary information for salient objections. In semantic segmentation and object detection tasks, the proposed fusion network, OCCO, also achieves outstanding detection performance.

\bmhead{Acknowledgements}
This work was supported by the National Natural Science Foundation of China (62202205), the National Key Research and Development Program of China under Grant (2023YFF1105102, 2023YFF1105105) and the Fundamental Research Funds for the Central Universities (JUSRP123030).










\section*{Declarations}


\begin{itemize}
\item \textbf{Conflict of interest}
The authors declare that they have no known competing financial interests or personal relationships that could have appeared to influence the work reported in this paper.

\item \textbf{Data availability }
Data will be made available on request.
\end{itemize}

\bibliography{bio_response}

\end{document}